\documentclass[sigconf]{acmart}
\renewcommand\footnotetextcopyrightpermission[1]{} 

\usepackage{xspace}
\usepackage{subfig}
\usepackage{tabu}
\usepackage{tabularx}
\usepackage{relsize}
\usepackage{float}
\usepackage{graphicx}
\usepackage{amsfonts}
\usepackage{amsmath}
\usepackage{enumitem}
\usepackage{adjustbox}
\usepackage{multirow}
\usepackage{bm}
\usepackage{listings}
\acmConference[KDD]{KDD}{2022}{USA}

\newif\ifshowcomments
\showcommentstrue

\ifshowcomments
\newcommand{\mynote}[2]{\fbox{\bfseries\sffamily\scriptsize{#1}}
{\small$\blacktriangleright$\textsf{\emph{#2}}$\blacktriangleleft$}}
\else
\newcommand{\mynote}[2]{}
\fi

\newcommand{\algo}{Fed-TGAN\xspace}
\newcommand{\alg}{Fed\xspace}
\newcommand{\central}{Centralized\xspace}

\newcommand{\vanilla}{vanilla FL-TGAN\xspace}
\newcommand{\mdtgan}{MD-TGAN\xspace}
\newcommand{\mdtg}{MD\xspace}

\newcommand{\p}{$P$\xspace}
\newcommand{\q}{$Q$\xspace}
\newcommand{\Ni}[1]{$N_{{#1}}$\xspace}
\newcommand{\N}{$N$\xspace}
\newcommand{\Dij}[1]{$D_{{#1}j}$\xspace}
\newcommand{\dcaij}{$X_{ij}$\xspace}
\newcommand{\dcaj}{$X_{j}$\xspace}
\newcommand{\lej}{$LE_{j}$\xspace}
\newcommand{\vgmij}[1]{$VGM_{{#1}j}$\xspace}
\newcommand{\vgmj}{$VGM_{j}$\xspace}

\newcolumntype{L}[1]{>{\raggedright\let\newline\\\arraybackslash\hspace{0pt}}m{#1}}
\newcolumntype{C}[1]{>{\centering\let\newline\\\arraybackslash\hspace{0pt}}m{#1}}
\newcolumntype{R}[1]{>{\raggedleft\let\newline\\\arraybackslash\hspace{0pt}}m{#1}}

\newif\ifshownotationtable
\shownotationtablefalse

\newif\ifnotdoubleblind
\notdoubleblindtrue

\setlist[itemize]{leftmargin=*}
\begin{document}

\title{\algo: Federated Learning Framework for Synthesizing Tabular Data}


\ifnotdoubleblind


  

  \author{Zilong Zhao}
\affiliation{%
  \institution{TU Delft}
  \city{Delft}
  \country{Netherlands}}
\email{z.zhao-8@tudelft.nl}  

\author{Robert Birke}
\affiliation{%
  \institution{ABB Corporate Research Switzerland}
  \city{D\"{a}ttwil}
  \country{Switzerland}}
\email{robert.birke@ch.abb.com}

\author{Aditya Kunar}
\affiliation{%
  \institution{Tu Delft}
  \city{Delft}
  \country{Netherlands}}
\email{A.Kunar@student.tudelft.nl}

\author{Lydia Y. Chen}
\affiliation{%
  \institution{TU Delft}
  \city{Delft}
  \country{Netherlands}}
\email{Lydiaychen@ieee.org}

\fi


\begin{abstract}

Generative Adversarial Networks (GANs) are typically trained to synthesize data, from images and more recently tabular data, under the assumption of directly accessible training data. 
Recently, federated learning (FL) is an emerging paradigm that features decentralized learning on client's local data with a privacy-preserving capability. 
And, while learning GANs to synthesize images on FL systems has just been demonstrated, it is unknown if GANs for tabular data can be learned from decentralized data sources. Moreover, it remains unclear which distributed architecture suits them best. Different from image GANs, state-of-the-art tabular GANs require prior knowledge on the data distribution of each (discrete and continuous) column to agree on a common encoding -- risking privacy guarantees.
In this paper, we propose \algo, the first Federated learning framework for Tabular GANs. To effectively learn a complex tabular GAN on non-identical participants, \algo designs two novel features: (i) a privacy-preserving multi-source feature encoding for model initialization; and (ii) table similarity aware weighting strategies to aggregate local models for countering data skew. We extensively evaluate the proposed \algo against variants of decentralized learning architectures on four widely used datasets. Results show that \algo accelerates training time per epoch up to 200\% compared to the alternative architectures, for both IID and Non-IID data. 
Overall, \algo not only stabilizes the training loss, but also achieves better statistical similarity and machine learning utility between generated and original data. Our code is released at \url{https://github.com/zhao-zilong/Fed-TGAN}.

\end{abstract}






\maketitle
\pagestyle{plain}

\section{Introduction}
\begin{figure*}[htb!]
	\begin{center}
		\includegraphics[width=0.99\textwidth]{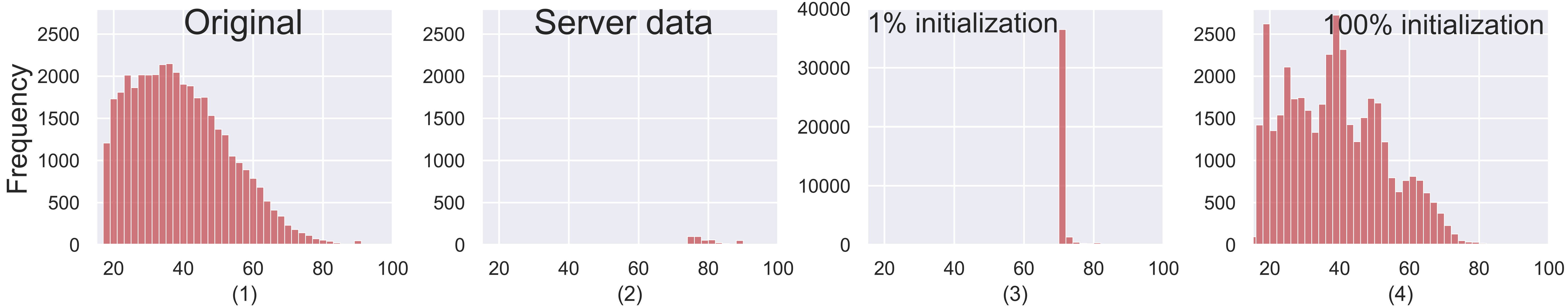}
		\vspace{-1em}
		\caption{Challenge of initializing column distribution of a tabular model with example of   \textit{age} column from the \textbf{Adult} dataset: (1) original data, (2) skewed 1\% sampled data to build VGM encoder, and (3) generated data based on model using VGM encoder built on skewed data. (4) Comparison with generated data based on model using VGM encoder built from original data.} 
		\label{fig:motivation}
	\end{center}
	\vspace{-.5em}
\end{figure*}

Generative Adversarial Networks (GANs)~\cite{gan} are an emerging methodology to synthesize data, ranging from images~\cite{stylegan, stylegan2}, to text~\cite{semeniuta2018accurate}, to tables~\cite{tablegan, ctgan}. The key components of GANs are training two competing neural networks, i.e., generator and discriminator, where the former iteratively generates synthetic data and the latter judges its quality. During the training process, the discriminator needs to access the original data and provide feedback to the generator by comparing it with the generated data. However, such a privilege of direct data access may no longer be taken for granted due to the ever increasing concern for data privacy. For instance, training a medical image generator~\cite{asyndgan} from multiple hospitals refrains from centralized data processing and calls for decentralized and privacy-preserving learning solutions.

In response to such a demand, the federated learning (FL) paradigm emerges.  FL features decentralized local processing, under which machine learning (ML) models can first be trained on clients' local data in parallel and subsequently be securely aggregated by the federator. As such, the local data is not directly accessed, except by the owner, and only intermediate model is shared. The key design choices of constructing a FL framework for GANs depends on how to effectively distribute the training of generator and discriminator networks across data sources. On the one hand, discriminators are typically located on clients' premises due to the need of processing the client's data. On the other hand, the prior art explores a disparate trend of training image generators: centrally at the server~\cite{mdgan} or locally at the clients~\cite{fegan}.  While tabular data is the most dominant data type in industries~\cite{arik2019tabnet}, there is no prior study on training GANs for tabular data under the FL paradigm. 

Training of state-of-the-art tabular GANs, e.g., TGAN~\cite{tgan}, CTGAN~\cite{ctgan} and CTAB-GAN~\cite{ctabgan}, from decentralized data sources in a privacy preserving manner presents multiple additional challenges as compared to image GANs. They are closely related to how current tabular GANs explicitly model each column, be it continuous or categorical variables, via data-dependent coding schemes and statistical distributions. Hence, the first challenge is to unify the encoding schemes across data sources that are non-identically independently distributed (Non-IID), and in a privacy preserving manner\footnote{Privacy preserving solutions refer to ones that do not require full knowledge of the local data.}. Secondly, the convergence speed of GANs critically depends on how to merge local models~\cite{jill_fed}. For image GANs~\cite{fedgan}, the merging weights are determined jointly by the data quantity and the (dis)similarity of class distribution across clients. Beyond that, tabular GANs need to consider a more fine-grained (dis)similarity mechanism for deciding merging weights, i.e., differences in every column across clients. 

In this paper, we aim to design a federated learning framework, \algo, that allows to train tabular GAN models from decentralized clients. The architecture of \algo is that (i) each client trains its  generator and discriminator networks using its local data and (ii) the federator aggregates the generators and discriminators. We also propose two algorithmic features that address more fine-grained per column modeling in a privacy preserving manner. First, the novel feature encoding scheme of \algo can reconstruct the entire column distribution via bootstrapping each client's partial information. Secondly, a more precise weighting scheme can effectively merge local models by considering the quantity and distribution dissimilarity for every column across all clients. 
We design and implement a first of its kind federated learning framework for tabular GANs using the PyTorch RPC framework.

We extensively evaluate \algo on a vast number of client scenarios, which have disparate data distributions. 
Specifically, \algo is compared with three architecture baselines: (1) centralized approach, (2) vanilla federated learning and (3) multiple discriminator architecture ~\cite{mdgan} comprising of a single generator and multiple discriminators. The evaluation is performed on four commonly used machine learning datasets where the statistical similarity between generated and real data are reported as evaluation metrics. Our results show that \algo remarkably reduces training time per epoch comparing to multi-discriminator solution by up to 200\%. 
Additionally, under an unbalanced amount of local data among the clients, \algo converges much faster than vanilla federated learning. And, for scenarios where data in all clients is non independently and identically distributed, the convergence of \algo is not only stable, but also provides better similarity between generated and real data. 
The main contributions of this study can be summarized as follows:
\begin{itemize}[leftmargin=*]

 \item We design and prototype a one-of-a-kind federated learning framework for the decentralized learning of tabular GANs (i.e. CTGAN) on distributed clients' data. We apply \algo with CTGAN as an example, but its usage is not limited to it.
 
\item We create a privacy preserving feature encoding method, which allows the federator to build global feature encoders (either for categorical or continuous columns) without accessing local data.

\item We design a table-similarity aware weighting scheme for merging local models, which is shown to achieve a faster convergence speed when the data quantity and data quality are highly imbalanced among clients.
 
\item We extensively evaluate \algo to synthesize four widely used tabular datasets on the prototype testbed. 
Across various clients scenarios, the \algo shows remarkably high similarity to the original data while also converging faster than vanilla FL and MD-GAN.
\end{itemize}

Our code is temporarily hosted on google drive\footnote{\url{https://drive.google.com/file/d/1ggjCrywVwk-8MG5Oq4TpYzohyqgz7cm3/view?usp=sharing}} with a detailed description for reproducing the results. It contains not only the training code, but also the evaluation scripts.

\section{Preliminary and Motivation}
\label{sec:motivation}

\begin{figure*}[t]
	\begin{center}
			\subfloat[Centralized GAN]{
			\includegraphics[width=0.24\textwidth]{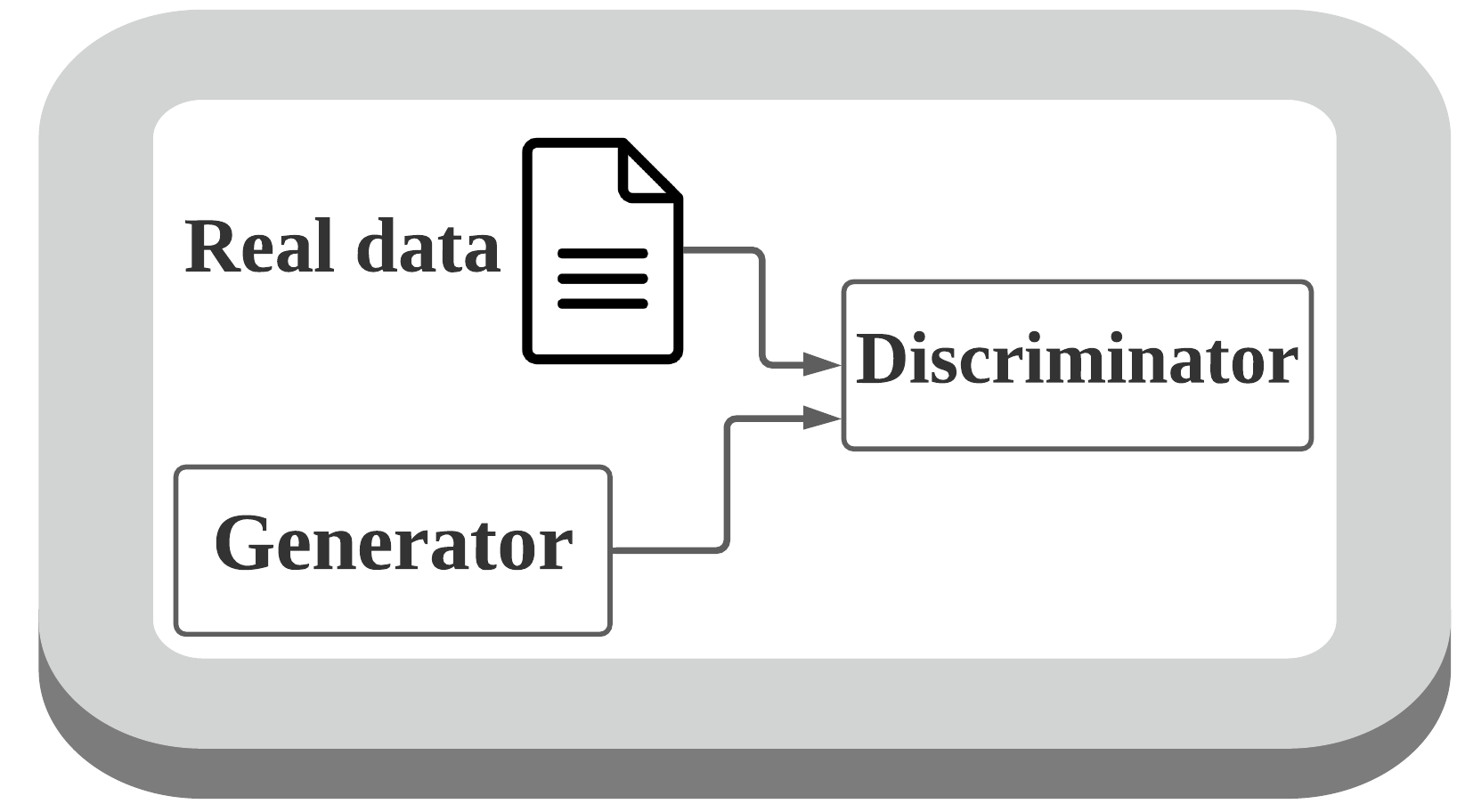}
			\label{fig:centralized_gan}
		}\hspace{2em}
		\subfloat[Multi-Discriminator structure]{
			\includegraphics[width=0.31\textwidth]{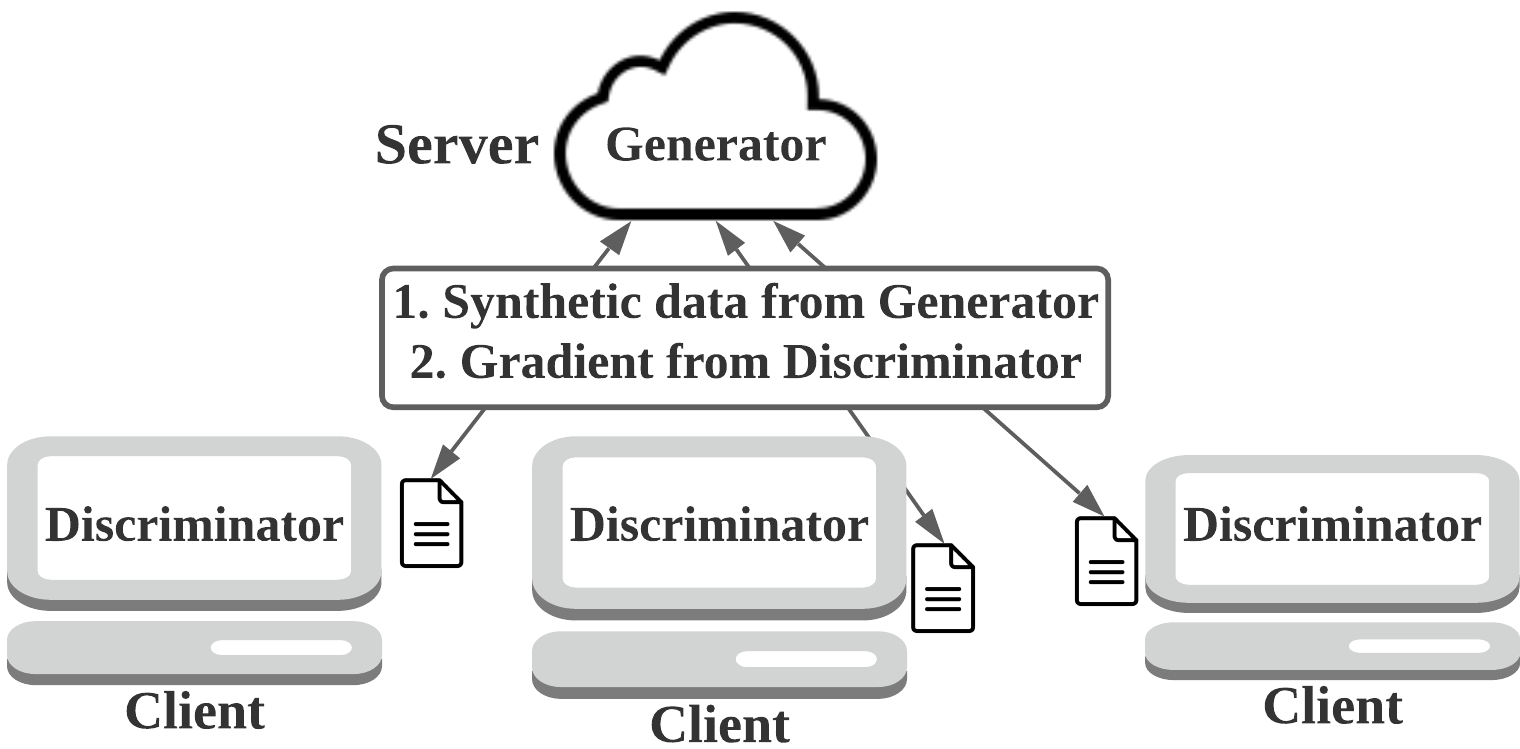}
			\label{fig:mdgan}
		}
		\hspace{2em}
		\subfloat[Federated Learning structure]{
			\includegraphics[width=0.31\textwidth]{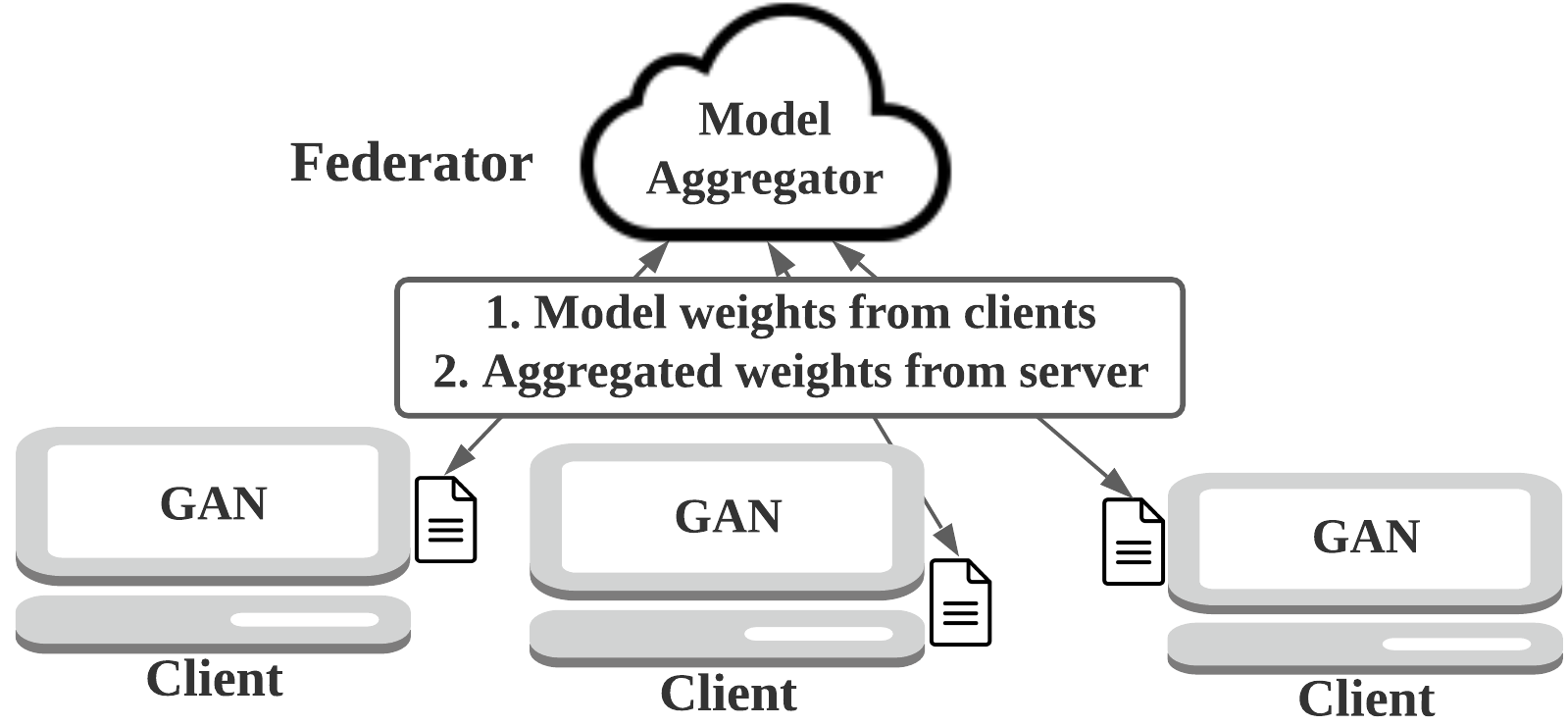}
			\label{fig:fedgan}
		}
		\caption{Privacy-preserving framework for distributed GAN}
		\label{fig:algo_structure_moti}
 	\end{center}
  	\vspace{-.5em}
\end{figure*}

{\textbf{Preliminary}}. Key in federated, and generally decentralized, learning is that all participating nodes use the same model structure. This structure heavily depends on the the input data type and its encoding. Previous federated GANs studies~\cite{mdgan, fedgan, asyndgan} focus on only one type of data i.e images. Image data makes it easy to pre-define the encoding and neural network structure independent from the specific images located at each participating node. However, the same does not apply to tabular data. Each columns requires an encoding which shapes the input layer influencing the model structure. These encodings depend on both the data type, i.e. categorical or continuous, and the data values, i.e. data distribution.  For example, the state-of-the-art generative tabular data models TGAN~\cite{tgan}, CTGAN~\cite{ctgan} and CTAB-GAN~\cite{ctabgan} use one-hot encoding for categorical columns and Variational Gaussian mixtures (VGM) encoding for continuous columns. Both encoding types require to know the per column global data properties. One-hot encoding requires the list of all possible distinct values and VGM encoding depends on the estimation of the all possible modes with their mean and variance. Hence image federated learning systems can not readily be applied to tabular data problems.

Tabular GANs on federated learning systems need to overcome one additional challenge with respect to traditional learning systems. To agree on a common encoding and, consequent, model structure it is important to know the column by column data distribution across all participants. This is straightforward if privacy is of no concern: e.g., collect all the client data on one node, decide the encoding and distribute the decision to all other nodes. However this goes against the fundamental aim of federated learning: training models without the need to share the detailed information of local data to preserve privacy. For categorical columns the problem can be solved by the participants sharing  the list of distinct values in each column with little to no privacy infringement. But for continuous columns the problem is not as straightforward due to the VGM requirement.



{\textbf{Motivation Example}}. We demonstrate the challenge of encoding continuous columns with the following experiment using the Adult dataset (see Sec.~\ref{ssec:setup} for details on the setup and dataset). Here we momentarily relax the privacy requirement and assume that the federator coordinating the federated learning has access to 1\% of the global data. This 1\% data is used to build the VGM encoders for all continuous columns which are then distributed and used by all clients to encode the local data. Without a global view it is impossible to know how well the 1\% data represents the global population. If this 1\% data is sampled in a skewed way it can severely degrade the encoding quality leading to poor model performance. We show this effect on the \textit{age} column. We select the 1\% data from tail of the age distribution. The distributions of the original and selected data are shown in the Fig.~\ref{fig:motivation}(1) and Fig.~\ref{fig:motivation}(2). Fitting a VGM encoder from the sampled data will encode well only the data between 75 and 90 years which, however, represents only the data above the 99$^{th}$ percentile of the real age distribution. Using this encoder to train a model leads to poor generation performance, i.e., the generated samples are not representative of the original data (see Fig.~\ref{fig:motivation}(3)). For comparison, the Fig.~\ref{fig:motivation}(4) shows the distribution of samples generated with a VGM encoder built from all of the data.

Another key issue for federated tabular learning is the weighting of models from different clients during model aggregation. This issue is exacerbated with tabular data. Non iid data across clients can lead to poor training convergence and ultimately model performance. Federated learning systems counter this effect by weighting each client model differently based on the similarity of local data to the global data. Image federated learning estimates this similarity based on the distribution of labels which can be seen as 1-dimensional data. But for tabular data, each column can be seen as one dimension requiring a multi-dimension solution. Moreover, while the same method as for image labels can be applied to categorical columns, one can not directly estimate the similarity for continuous columns without knowing all the data points. Thus, a new weighting method is needed for tabular data.

\begin{figure*}[t]
	\begin{center}
		\subfloat[Collect distributions, create encoders]{
			\includegraphics[width=0.3\textwidth]{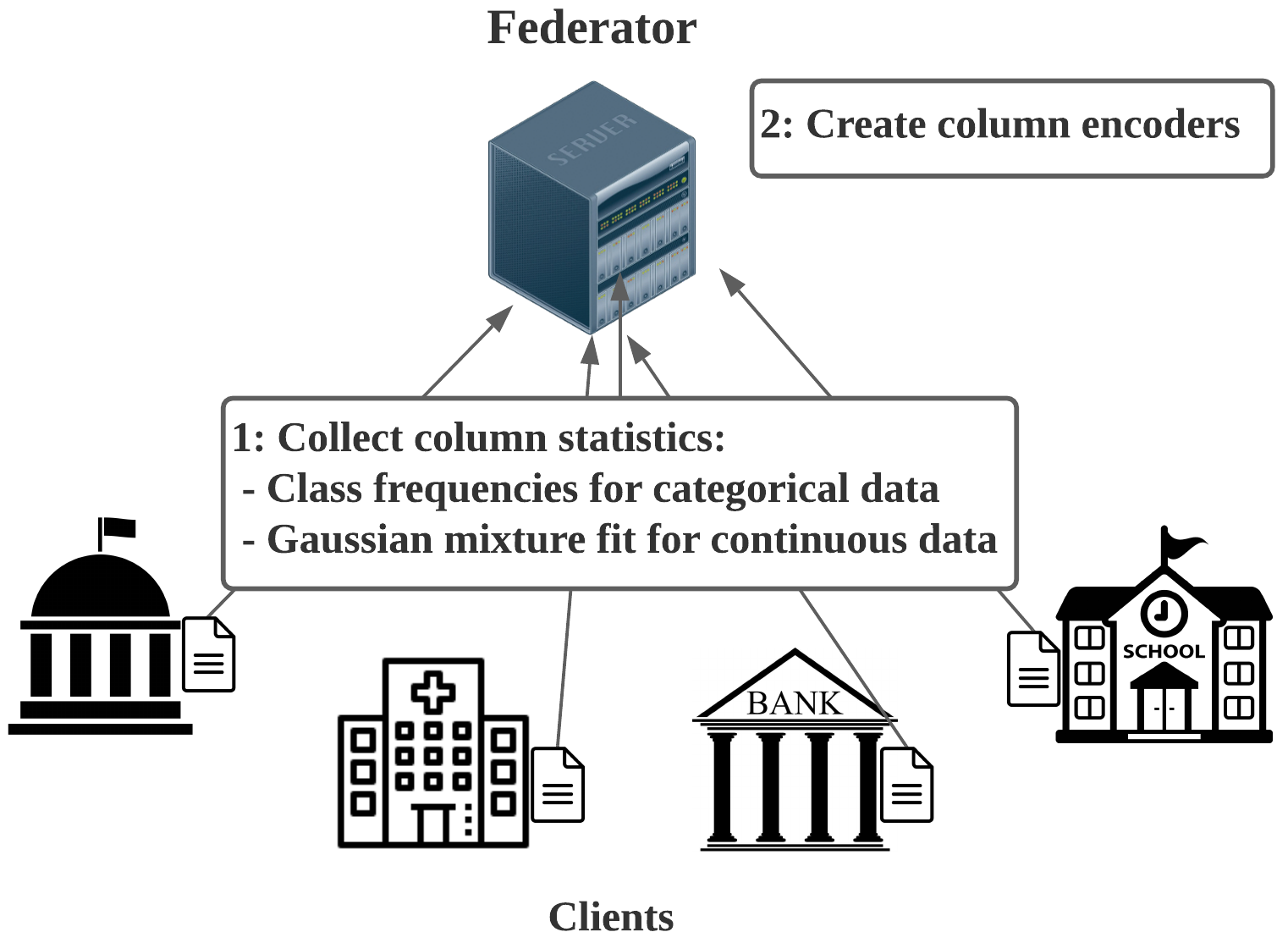}
			\label{fig:collect_distribution}
		}
		\hspace{2em}
		\subfloat[Distribute encoders, weight clients]{
			\includegraphics[width=0.3\textwidth]{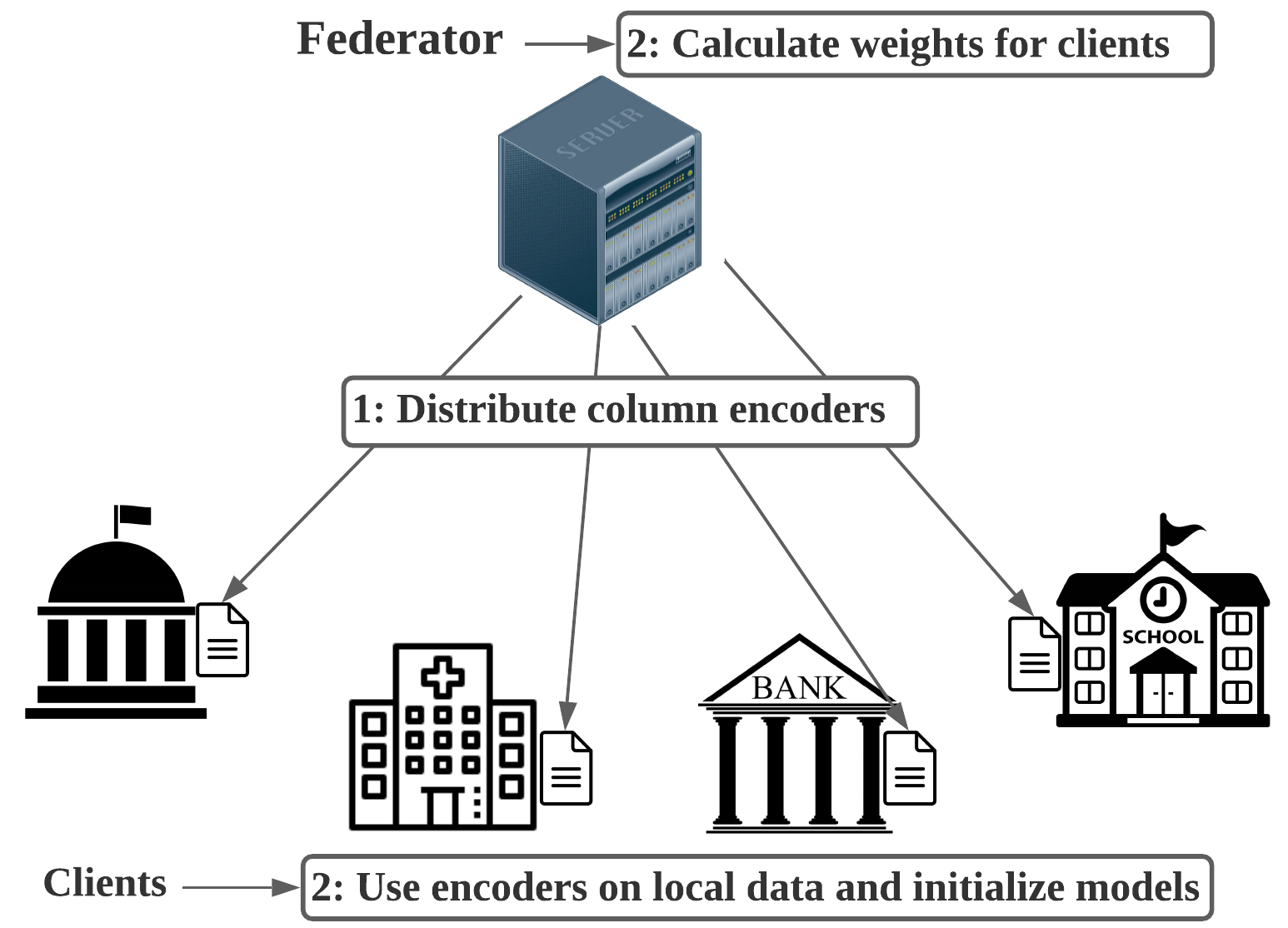}
			\label{fig:distribute_information}
		}
		\hspace{2em}
		\subfloat[Training]{
			\includegraphics[width=0.29\textwidth]{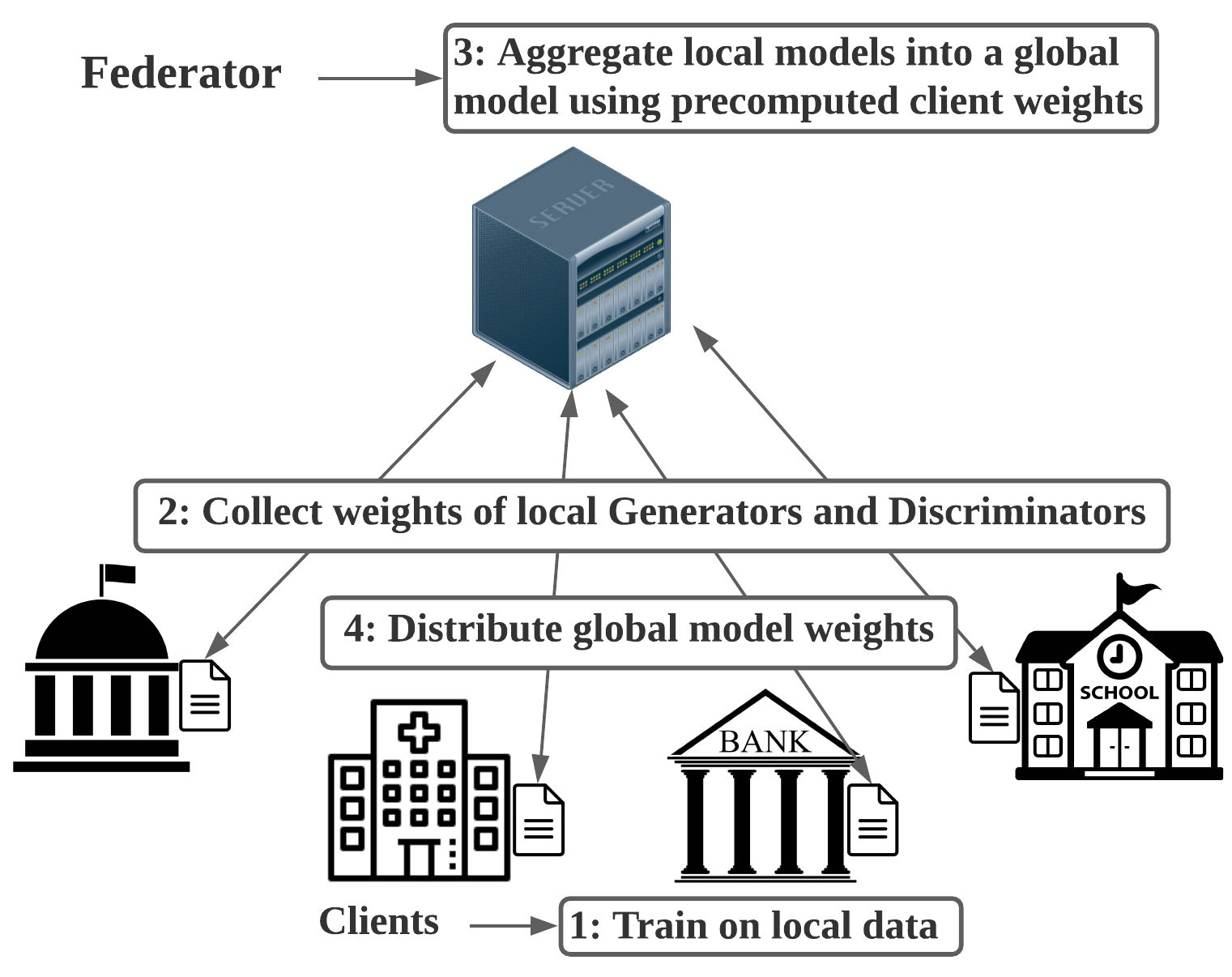}
			\label{fig:training_process}
		}
		\caption{Initialization and training process of \algo } 
		\label{fig:algo_structure_2}
 	\end{center}
  	\vspace{-.5em}
\end{figure*}

\section{Decentralized Architecture}
\label{sec:relatedwork}

GANs comprise two types of antagonizing deep neural networks: generator and discriminator. In training the two take turns. First the generator tries to synthesize data which as indistinguishable as possible from the real data to fool the discriminator. Then the discriminator tries to detect the fake from real data to counter the generator. 
Model weights are updated using a minimax loss function which tries to minimize the loss of the generator, related to the number of correctly detected fakes by the discriminator, while maximizing the loss of the discriminator, related to how often the generator successfully fooled the discriminator, see Fig.~\ref{fig:centralized_gan}. 

Training GANs is computationally demanding since they comprise of at least two deep neural networks (i.e., one generator and one discriminator) and are trained using big datasets. Therefore, a decentralized training framework can be highly beneficial in such a setting and is explored for image GANs only. Existing 
solutions to decentralized GANs training can be classified into two categories: (1) Multi-Discriminator (MD) structure~\cite{mdgan, asyndgan, temporary_dis} and (2) Federated Learning (FL) structure~\cite{fedgan, fegan, fl}. 

{\bf Multi-Discriminator} has one single generator in the server and multiple discriminators distributed across the clients. The structure is illustrated in Fig.~\ref{fig:mdgan}. 
The server determines the network architecture for the generator and discriminators. 
The generator is located at the sever and trains its network using random inputs and the gradients from all discriminators, as typically done for the centralized GANs. On the contrary, the discriminators located at the decentralized data sources train their networks locally using outputs from the generator, i.e., synthesized data. Such a structure ensures that the client's data does not need to leave the clients' machines.
The downside of the MD structure is that it induces significant communication overhead between the generator and discriminator, i.e., sending synthesized data to all discriminators, and returning discriminator's gradients to the generator per training epoch. 
In addition, client discriminators tend to over-fit to their local data with more training epochs. MD-GAN~\cite{mdgan} counters the latter issue by allowing clients to randomly swap their models in a peer-to-peer way every several epochs. Even so, each discriminator is treated with the same weight to update the generator. Thus, the convergence of the generator is not optimal~\cite{fegan} when the quantity and distribution of data is highly skewed among clients.



{\bf Federated Learning} \cite{fl} structure (shown in Fig.~\ref{fig:fedgan}) is composed of multiple GANs (with a discriminator and generator) on each client who has direct access to data. Each client first trains a GAN using the local data. Then sends the GAN model to the federator. The key role for the federator is: (i) during initialization to determine the GANs architecture; and (ii) during training to aggregate the local GANs models into a global GAN and redistribute it to all clients. 
Communication occurs when clients upload their model weights to the federator and when the federator redistributes the updated weights. Such a communication and merging local models is commonly refereed a global {\emph{training round}} in FL studies. The resulting overhead is lower than for the MD structure that requires communication between server and clients per training epoch. Additionally,  transferring model weights to/from the federator is more efficient than transferring synthesized data to each discriminator in the case of the MD structure.
The FL structure also has a strong scalability relative to the number of clients, as the computation complexity of model aggregation is lower than training the generator network. 
Another advantage of FL structure is to allow weighting local models during aggregation, which helps to accelerate the convergence of the generator under skewed data distributions among clients.   
Local data ratios and Kullback-Leibler (KL) weighting methods from~\cite{fegan} are introduced to address skewed data challenges for image data.

\noindent\textbf{Architecture choice for \algo}. The FL structure has multiple benefits, ranging from communication overhead, scalability, training stability, and handling skewed client data, compared to the MD structure. In this work, we thus adopt the FL structure for enabling training tabular GANs on decentralized data sources. In summary, \emph{the proposed \algo is composed of one federator and multiple clients, following  the training procedure of the FL structure.}

\section{\algo}
\label{sec:model}
In this section we introduce the design of \algo which adapts the FL structure presented in Sec.~\ref{sec:relatedwork} to overcome the challenges presented in Sec.~\ref{sec:motivation}. First, We add an initialization step to standardize the encoding for each column across all participants. Second, we choose the best encoding in a privacy-preserving manner by estimating the global data distribution without directly accessing the participant's local data. Third, we introduce a multi-dimensional weighting mechanism to ensure model convergence under Non-IID data distributions across multiple columns. 

\subsection{Privacy-preserving feature encoding}
Our privacy-preserving model initialization comprises two steps as shown in Fig.~\ref{fig:collect_distribution} and~\ref{fig:distribute_information}.

{\bf Step 1}. Each of the \p clients extracts the statistical properties of the local data and sends them to the federator. The information sent is different based on the column type. For any categorical column $j$, each client $i$ computes and sends in the category frequency distribution \dcaij. This information is used in three ways. First, the federator uses all distinct categories to build the label encoder \lej for column~$j$.
A label encoder is a table which maps all possible distinct values of a categorical column into their corresponding rank in one-hot encoding. Second, the frequency information is used to build an aggregated global frequency distribution \dcaj for column $j$. Third, the sum of the frequency values is used to compute the number of table rows: i) available locally \Ni{i} at each client $i$; ii) available globally \N across all clients.
The global label frequency distribution \dcaj, \Ni{i} and \N are needed to estimate the similarity of clients' local data for computing the clients' weights for model aggregation. If no categorical columns are present in the tabular data, the client sends out \Ni{i} instead.

For any continuous column $j$, each client $i$ fits and sends in the parameters of a VGM model \vgmij{i}. To estimate the global distribution of column $j$ the federator uses \vgmij{1}, \vgmij{2}, $\dots$, \vgmij{P} to create the data sets \Dij{1}, \Dij{2}, $\dots$, \Dij{P} with \Ni{1}, \Ni{2}, $\dots$,\Ni{P} data points where $P$ is the number of clients. The federator then uses these data sets to fit a new global VGM model \vgmj for column $j$\footnote{It might be possible to fit the global model directly from the parameters of the local models by, e.g., adapting \cite{DBLP:conf/icpr/BruneauGP08}. This is left for future work.}.
\vgmj is used as the final encoder for column $j$. 


{\bf Step 2}. The federator distributes all the column encoders \lej and \vgmj to each client. Clients use this information to encode the local data and initialize the local models. Models initialized by using the same encoders will have the same input/output layers. This solves the first challenge outlined in Sec.~\ref{sec:motivation}. Note that the number and structure of the internal layers used for the generator and discriminator networks are predefined and independent of the data. In our evaluation against the MD structure this information is also used by the server to initialize the hosted generator network.
Note that in this process the federator never directly accesses the local data of the clients, only their statistical distribution, thus this addresses the second challenge from Sec.~\ref{sec:motivation}. 


\subsection{Table-ware similarity weighting scheme}

\begin{figure*}[t]
	\begin{center}
			\includegraphics[width=0.99\textwidth]{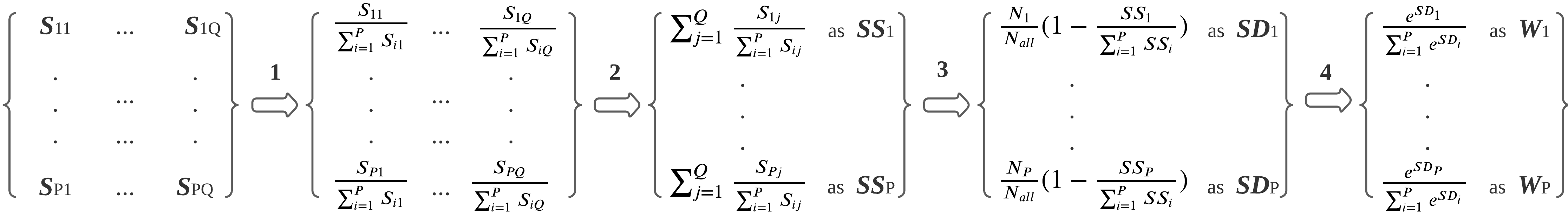}
		\caption{Weights calculation of \algo. Starting with a matrix of divergence scores for each client and table: (1) normalizes scores by column; (2) aggregates the scores per client; (3) incorporates differences in local data quantities; and, (4) performs the final weight normalization.} 
		\label{fig:algo_weights}
 	\end{center}
  	\vspace{-.6em}
\end{figure*}

After model initialization, federator uses the collected global data statistics to pre-compute the weights for each client. These weights are used in training during the model aggregation (shown in Fig.~\ref{fig:training_process}) to smooth convergence in the presence of skewed data across the clients. The weights calculation process is presented in Fig.~\ref{fig:algo_weights}.

{\bf Step 0} is to build a \p$\times$\q divergence matrix $\bm{S}$ where \p is the number of clients and \q is the number of columns. 
Each matrix element $S_{ij}$ is the divergence of client $i$ for column $j$ when compared to the global statistics of column $j$. The metric used depends on the type of column.

    {\bf Categorical columns} use the Jensen-Shannon Divergence (JSD)~\cite{jsd}. The JSD between two probability vectors $p$ and $q$ is defined mathematically as $\sqrt{\frac{D(p||m)+D(q||m)}{2}}$ where $m$ is the point-wise mean of $p$ and $q$, and $D$ is the Kullback-Leibler divergence~\cite{Joyce2011}. The JSD distance metric is symmetric and bounded between 0 and 1 enabling a hassle-free interpretation of results. For each categorical column $j$ and client $i$ we compute $S_{ij}$ as JSD between \dcaij and \dcaj,  i.e., $S_{ij} = \text{JSD}($\dcaij, \dcaj$)$.

    {\bf Continuous columns} use the Wasserstein Distance (WD)~\cite{wgan_test}. The first Wasserstein distance between two distributions $u$ and $v$ is defined as $WD(u,v)=inf_{\pi\in\Gamma(u,v)}\int_{\mathbb{R}\times\mathbb{R}}|x-y|d\pi(x,y)$ where $\Gamma(u,v)$ is the set of probability distributions on $\mathbb{R}\times\mathbb{R}$ whose marginals are $u$ and $v$ on the first and second factors, respectively. It can be interpreted as the minimum cost to transform one distribution into another where the cost is given by amount of distribution to shift times the distance it must be shifted.
    For each continuous column $j$, we use the data sets \Dij{i} created previously for each client $i$ to compute $S_{ij}$ as the WD between \vgmij{i} and \vgmj.

{\bf Step 1} normalizes the matrix $\bm{S}$ across the \p clients for each table column $j$. This is done by dividing each matrix element by the sum of the elements in the corresponding matrix column. This step maintains the relative divergence between different clients with respect to the global column data distribution while allowing to give the same importance to all columns (all columns sum up to 1).

{\bf Step 2} aggregates the divergence across the different table columns $j$. This is done via a sum along the rows of the matrix. For each client $i$ the resulting score $SS_i$ can already represent the divergence between client and global data distribution, but it does not yet take into account possible difference in the amount of local data available at each client.

{\bf Step 3} fuses the divergence in data values and data quantity at each client. Step 3 first normalizes the divergence metric between 0 and 1 across the clients. Then it uses the complement to represent similarity instead of divergence and combines it with the ratio of local data available with respect to the global data, i.e., $\frac{N_i}{N_{all}}$. The resulting $SD_{i}$ take into account differences in both number of values and distribution of values of the local vs. global data. It takes into account all different dimensions given by the different columns addressing the third challenge from Sec.~\ref{sec:motivation}.

{\bf Step 4} computes the final weights $W_i$. The $W_i$ for each client $i$ are obtained by passing the $SD_{i}$ to a softmax function. $W_i$ is the weight that the federator will use when it aggregates the model from  client $i$.

It is worth to mention that Fr\'echet Inception Distance (FID)~\cite{fid} is a common metric to measure the dissimilarity between real and synthetic images. 
FID considers the mean and the covariance of the data which works for continuous values (e.g., pixels of images) but not for categorical values. This makes FID not suitable for tabular data.


\subsection{Implementation details}

\algo is implemented using the Pytorch RPC framework.
This choice makes it easy to control the flow of the training steps from the federator. Clients just need to join the group, then wait to be initialized and assigned work. To parallelize the training across all clients, RPC provides a function \textit{rpc\_async()} which allows the federator to make non\-blocking RPC calls to run functions at a client.
To implement synchronization points, RPC provides a blocking function \textit{wait()} for the return from previously called function \textit{rpc\_async()}. 
The return of \textit{rpc\_async()} is \textit{future} type object. Once the \textit{wait()} is called on this object, the process is blocked until the return values are received from the client.
The federator starts the training on all clients via \textit{rpc\_async()}. Then it waits for the new model from each client via the \textit{wait()}. Once all models are received, they are aggregated into a single model using the clients weight and redistributed to the clients via \textit{rpc\_async()}. Once all clients confirm the reception of the updated model (via \textit{wait()}) the federator starts the next round of training. We plan to provide the source code via Github after publication of the paper.

One weakness of current RPC framework from Pytorch v1.8.1 is that it does not support the transmission of tensors directly on GPU through RPC call. This means that each time when we collect or update the model weights we need to pay an extra time cost to detach the weights from GPU to CPU or reload the weights from CPU to GPU.  
\section{Experimental Analysis}
\label{ssec:setup}


Our algorithm \algo is evaluated on four commonly used datasets, and compared with three alternative architectures. To evaluate the similarity between real and synthetically generated data, we resort to use avg-JSD  and avg-WD for categorical columns and continuous columns, respectively. We also provide an ablation analysis to highlight the efficacy of the proposed client weighting strategies of \algo. A training time analysis is reported in the end, to show the time efficiency of all algorithms.

\begin{table}[htb]
\centering
\caption{Description of datasets.}
\resizebox{0.7\columnwidth}{!}{
\begin{tabular}{ |c|c|c|c|c| }
\hline
\multirow{2}{*}{\textbf{Dataset}} & \multirow{2}{*}{\textbf{Rows [\#]}} & \multicolumn{3}{c|}{\textbf{Columns [\#]}}\\
\cline{3-5}
& & \textbf{Categorical}  & \textbf{Continuous}& \textbf{Total} \\
\hline
Adult     & 40k   & 9   & 5 & 14 \\
Covertype & 40k   & 45   & 10 & 55 \\
Credit    & 40k   & 1  & 30  & 31\\
Intrusion & 40k   & 20   & 22 & 42  \\
\hline
\end{tabular}
}
\label{table:DD}
 \vspace{-1em}
\end{table}
\subsection{Experimental setup}

\hspace{1em} {\bf  Datasets}. We test our algorithm on four commonly used machine learning datasets. {Adult}, 
{Covertype} and {Intrusion} -- are from the UCI machine learning repository~\cite{UCIdataset}, and {Credit} is from Kaggle~\cite{kagglecredit}. Due to our computational limitation, we randomly sample 40k data from each of above datasets. The details of each dataset are shown in Tab.~\ref{table:DD}. 



{\bf Baselines}. We compare \algo against 3 baselines: (i) multi-discriminator structure, (ii) {vanilla federated learning structure} and (iii) {centralized approach} abbreviated as \textbf{MD-TGAN}, \textbf{vanilla FL-TGAN}, and \textbf{Centralized}, respectively. The aim is to learn a CTGAN model from distributed clients using the three frameworks on the basis of CTGAN's default settings for encoding features~\cite{ctgan}. Specifically, we use 10 as the limiting number for estimated modes for the VGM encoders for each continuous column and use one-hot-encoding for categorical columns. We re-implement all baselines using the Pytorch v1.8.1 RPC framework. 

MD-TGAN clients swap discriminator models with each other at the end of each training epoch~\cite{mdgan}. For a fair comparison with MD-TGAN, we force also \algo and vanilla FL-TGAN to share the model weights with the federator at the end of each training epoch. Due to this, the notion \textit{per round} commonly referred in FL studies equals to \textit{per epoch} in this paper, if not otherwise stated.
Vanilla FL-TGAN is identical to \algo, except it uses identical weights for all clients equal to $\frac{1}{P}$ where $P$ is the number of clients.
Due to the different learning speed per epoch of the four frameworks, for a fair comparison we fix the number of epochs so that the training time is similar. In particular we use
to 500, 500, 500, and 150 epochs for \algo, Vanilla FL-TGAN, centralised, and MD-TGAN, respectively. 
We repeat each experiment 3 times and report the average.

{\bf Testbed}. Experiments are run under Ubuntu 20.04 on two machines. Each machine is equipped with 32 GB memory, GeForce RTX 2080 Ti GPU and 10-core Intel i9 CPU. Each  CPU core has two threads, hence each machine contains 20 logical CPU cores in total. The machine are interconnected via 1G Ethernet links (measured speed: 943Mb/s). One machine hosts the federator, the other all the clients. When not otherwise stated both federator and clients use the GPU for training. For experiments in Sec.~\ref{ssec:training_time}, when CPU is used to host clients for \algo and MD-TGAN, CPU affinity (by \textit{taskset} command in Linux) is used to bind each client to one logical CPU core to reduce interference between different processes.



    

    
\begin{figure}[t]
	\begin{center}
			\includegraphics[width=0.8\columnwidth]{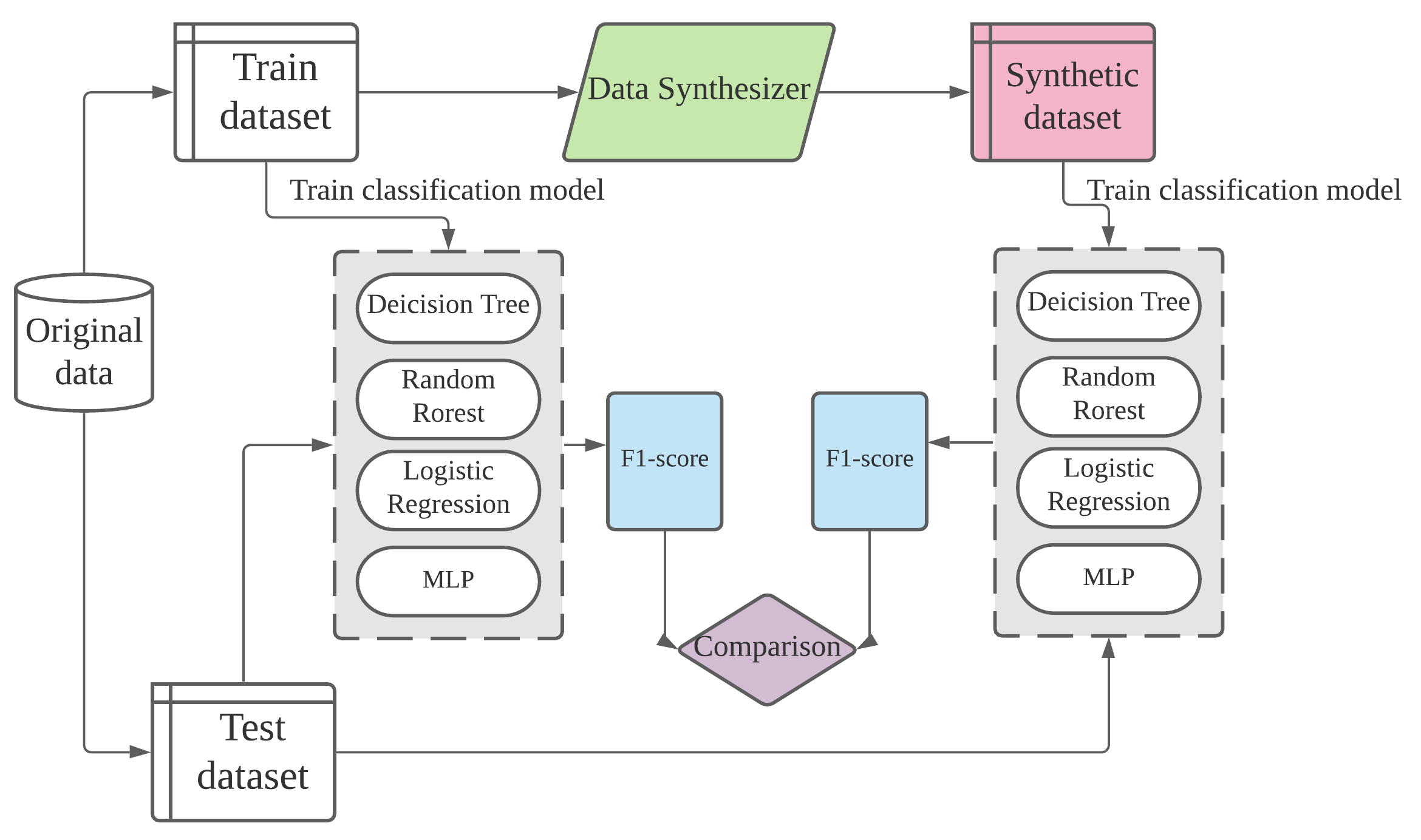}
		\caption{Evaluation flow for ML utility} 
		\label{fig:testing_flow}
 	\end{center}
   	\vspace{-1.6em}
\end{figure}

\subsection{Evaluation metrics}
\subsubsection{Statistical similarity}
\label{sec:metrics}
To evaluate the performance of the generator, a 40k synthetic data is sampled from the trained generator at the end of each epoch. We use two metrics to 
measure the statistical similarity between the real and synthetic data:
    
    {\bf Average Jensen-Shannon divergence (Avg-JSD).} Used for categorical columns (see Sec.~\ref{sec:model} for its definition). First, we compute the JSD  between the synthetic and real data for each categorical column. Second, we average the obtained JSDs to obtain a compact comprehensible score, abbreviated as {Avg-JSD}. The closer to 0 {Avg-JSD} is, the more realistic the synthetic data is.



{\bf Average Wasserstein distance (Avg-WD).} Used for continuous columns\footnote{We use WD over JSD for continuous columns since JSD is not well-defined when the synthetic values lie outside of the original value range from the real dataset, i.e., the KL divergence is not defined when comparing the similarity of probability distributions with non-overlapping support.} (see Sec.~\ref{sec:model} for its definition). Unlike JSD, WD is unbounded and can vary greatly depending on the scale of the data. To make the WD scores comparable across columns, before computing the WD we fit and apply a min-max normalizer to each continuous column in the real data and apply the same normalizer to the corresponding columns in the synthetic data. We average all column WD scores to obtain the final score abbreviated as {Avg-WD}. The closer to 0 {Avg-WD} is, the more realistic the synthetic data is.

\subsubsection{Machine learning utility} To quantify the ML utility between the synthesized and real data we use the difference in F1-score. The F1-score captures both the precision and recall of a ML model.

{\bf F1-score difference.} We compare the performance achieved by 4 widely used ML algorithms: decision tree, random forest, logistic regression and MLP. We implement the experiments using Python and scikit-learn 0.24.2. We use the default value for all hyper parameters except:
the max-depth for decision tree and random forest models is set to 38 and MLP uses one 128-neuron hidden layer. The evaluation flow is shown in Fig.~\ref{fig:testing_flow}. The original data is split into a \textit{train dataset} (i.e., the datasets described in Tab.~\ref{table:DD}) and \textit{test dataset}. We use the train datasets to train our \algo and synthesize a dataset of the same size. Then we use again the \textit{train dataset} and the \textit{synthetic dataset} to train two separate instances of the 4 ML models listed above. The ML utility is measured via difference in F1-score between each models pair. The aim of this design is to test how close the ML utility is when we train a machine learning model using the synthetic data versus using the real data. We report the obtained differences in the results tables. 

\begin{figure}[t]
	\begin{center}
		\subfloat[Avg-JSD by epoch]{
			\includegraphics[width=0.46\columnwidth]{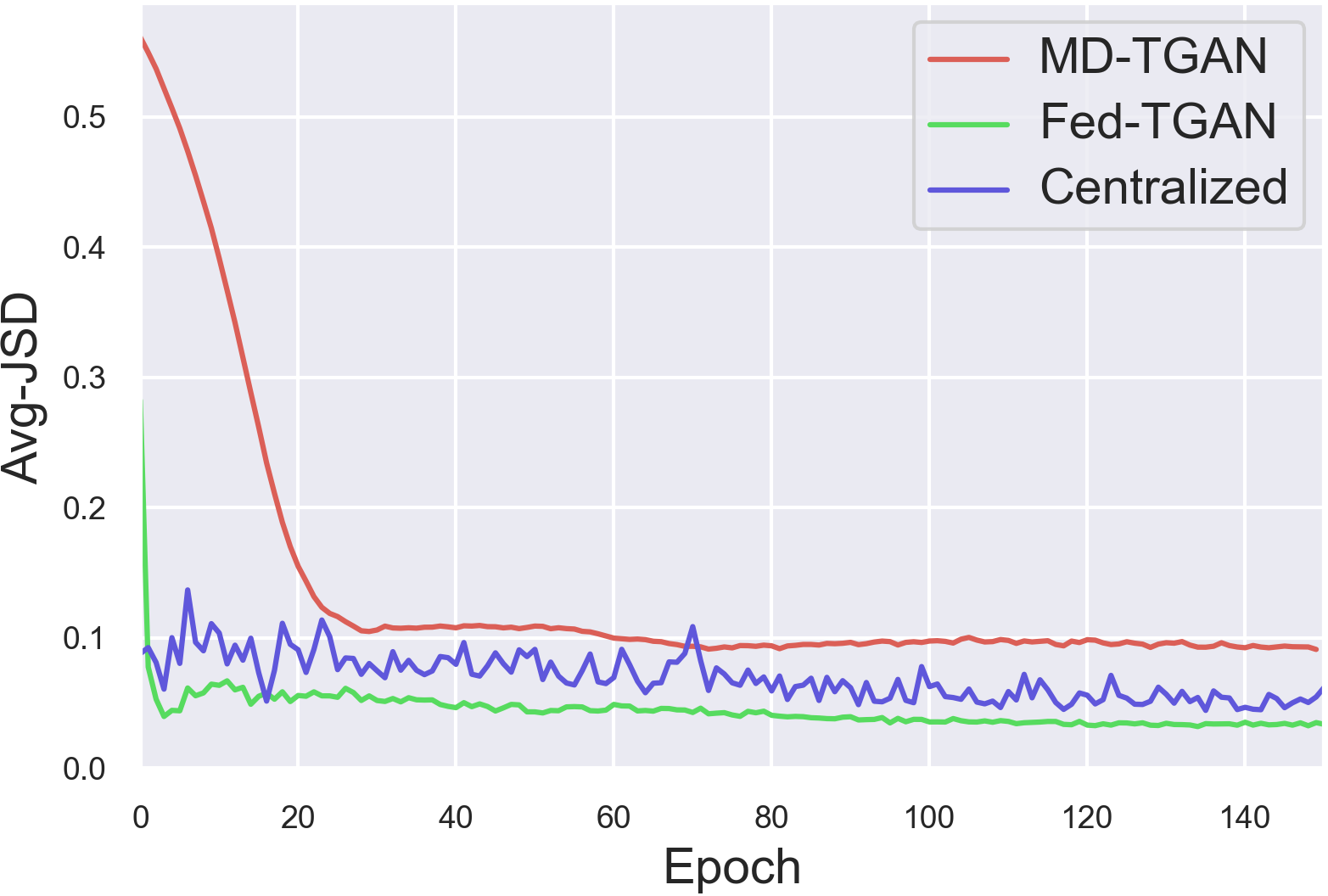}
			\label{fig:5full_intrusionjsd}
		}
		\subfloat[Avg-JSD by time]{
			\includegraphics[width=0.46\columnwidth]{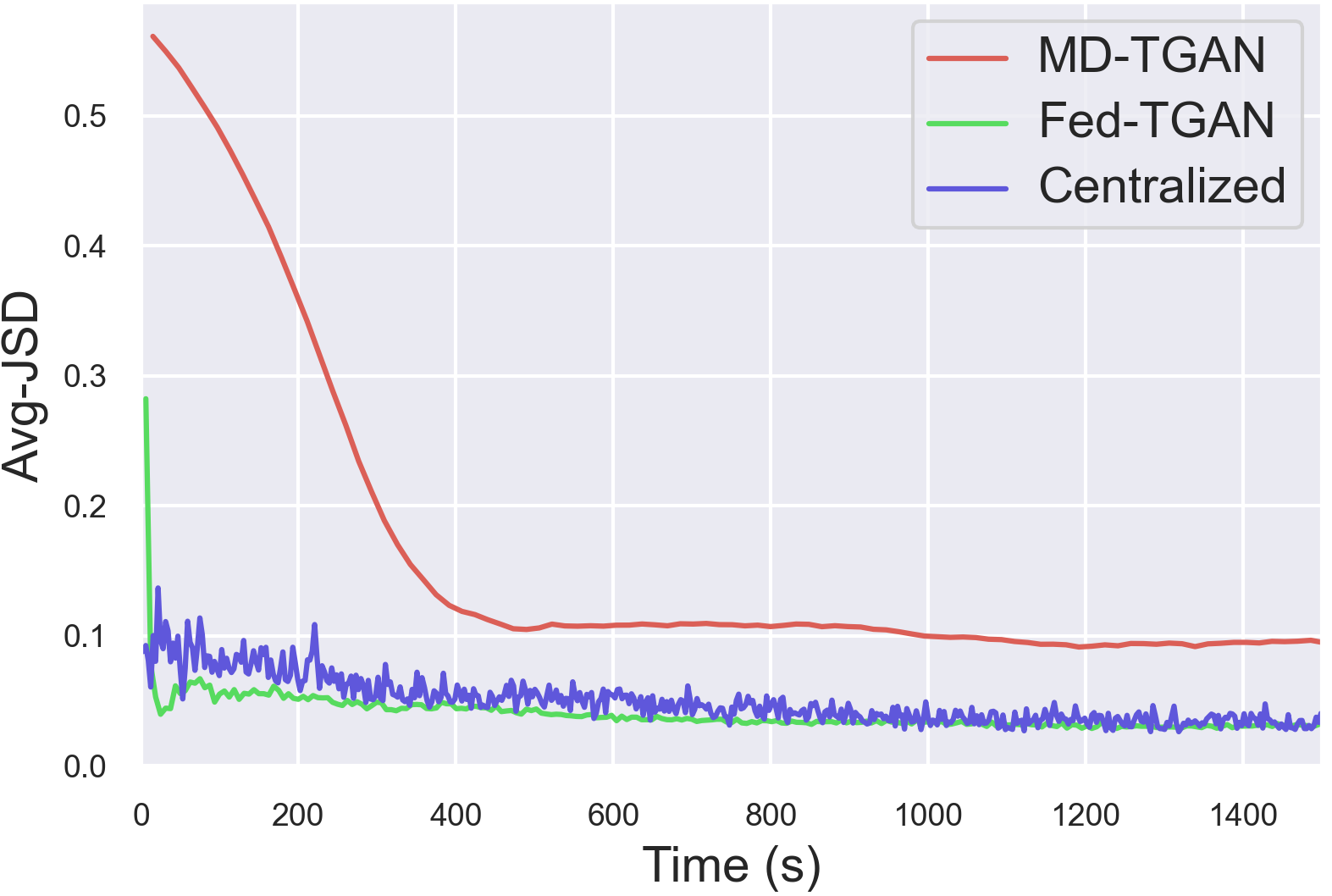}
			\label{fig:5full_intrusionjsdt}
		}
		\\[-2ex]
	    \subfloat[Avg-WD by epoch]{
			\includegraphics[width=0.46\columnwidth]{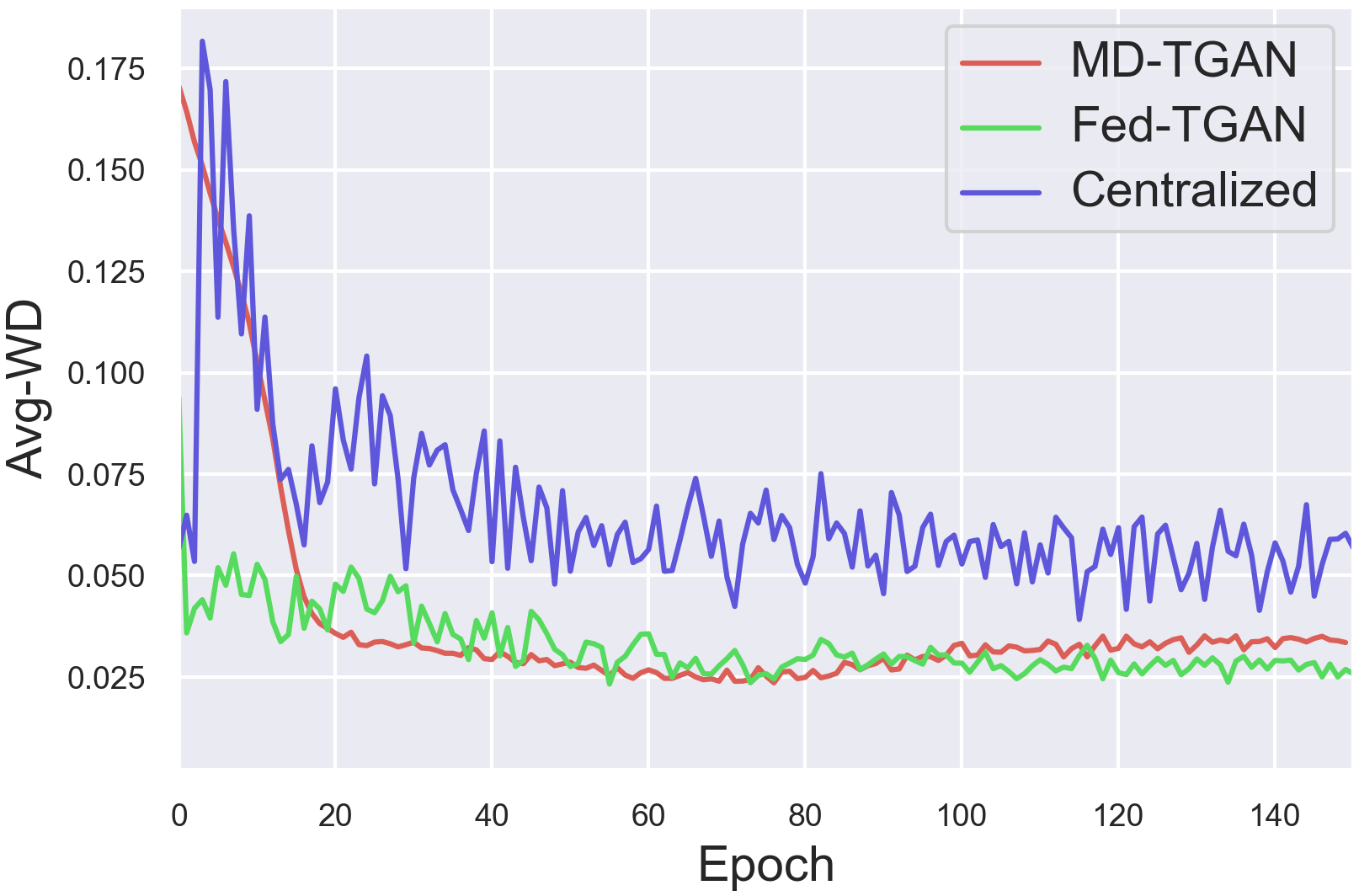}
			\label{fig:5full_intrusionwd}
		}
		\subfloat[Avg-WD by time]{
			\includegraphics[width=0.46\columnwidth]{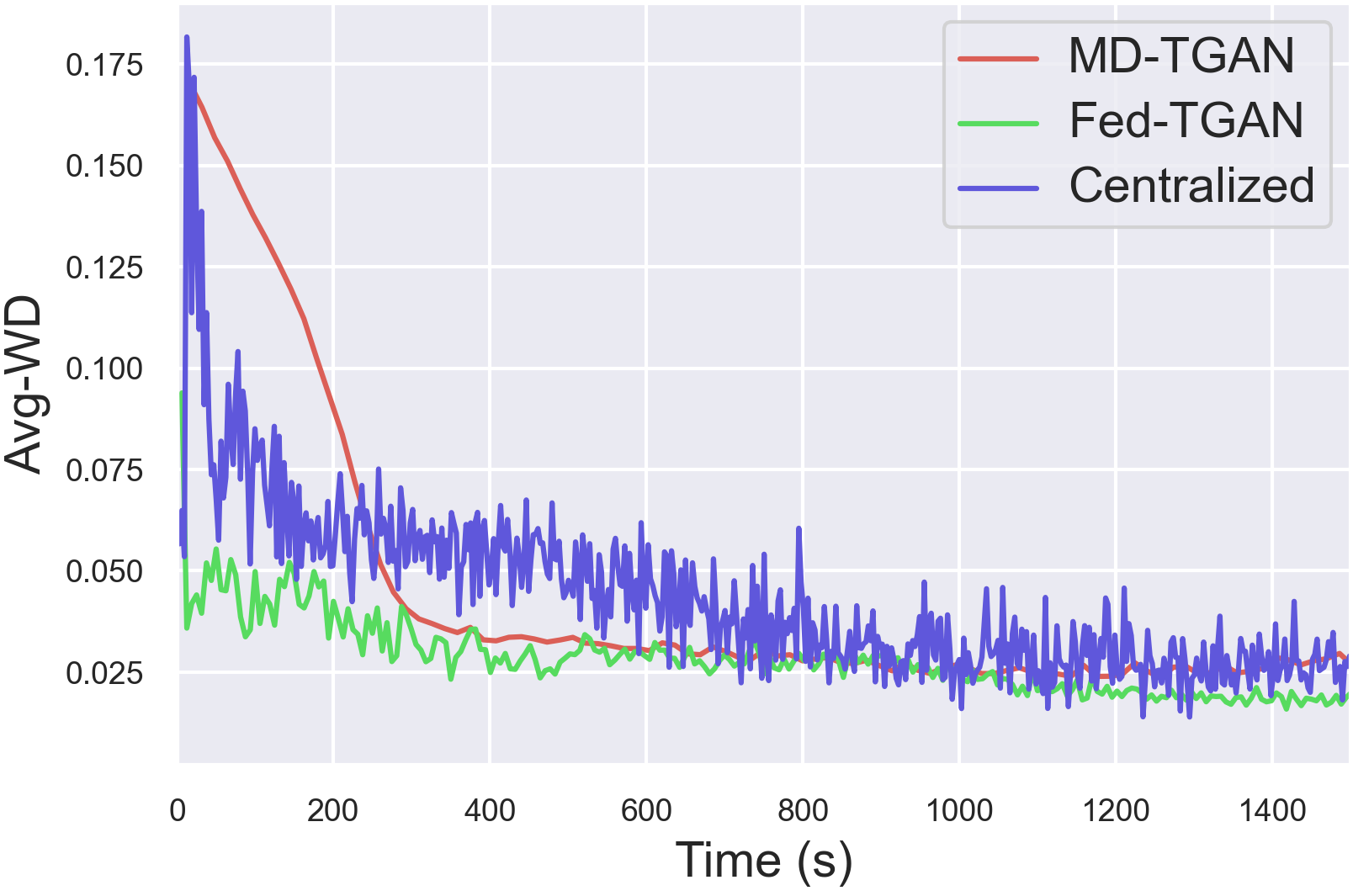}
			\label{fig:5full_intrusionwdt}
		}
		\caption{\mdtgan, \algo and \central: 5 clients each with a  \textit{complete} Intrusion data copy.}
		\label{fig:5full_intrusion}
	\end{center}
 	\vspace{-1.2em}
\end{figure}

\subsection{Result analysis}
We first evaluate how realistic the generated synthetic data is. Sec.~\ref{ssec:iid} designs an experiment where all the clients contain the whole dataset, this is to test the performance of each framework under the ideal case. Then in Sec.~\ref{ssec:imbalanced}, we implement a scenario where data on clients are IID, but quantities of data are highly imbalanced across all the clients. The objective of this experiment is to show the effect of our model aggregation weighting method. In the end in Sec.~\ref{ssec:ablation}, an ablation analysis is designed where one of the clients has much higher amount of data, but the data quality is low.  We want to show the efficacy of our table-similarity aware weighting method in the calculation of model aggregation weights.

\subsubsection{Ideal case of full dataset}
\label{ssec:iid}

This experiments uses one server (or federator) and 5 clients. Each client is provided with a copy of the full real dataset. This represents the ideal case with perfectly identical clients, i.e. each client has identical IID data. We compare in particular \algo, \mdtgan and \central. Since in this case the aggregation weights of \algo are the same as for \vanilla  due to the identical data, we skip \vanilla. Results for the Intrusion dataset are shown in Fig.~\ref{fig:5full_intrusion}. Avg-JSD and Avg-WD are presented by both epoch and time (in seconds) as different architectures spend vastly different time per epoch. For categorical columns \algo converges faster both by epoch and by time (see Fig.\ref{fig:5full_intrusionjsd} and \ref{fig:5full_intrusionjsdt}). Moreover, the Avg-JSD of \mdtgan converges quite slowly after epoch 24. For continuous columns, from the perspective of number of epochs, Avg-WD for \algo converges faster at the beginning, then becomes slightly worse than the Avg-WD for \mdtgan (see Fig.~\ref{fig:5full_intrusionwd}). However, inspecting the result by time, \algo not only converges faster, but also achieves a lower Avg-WD than the other two architectures (see Fig.~\ref{fig:5full_intrusionwdt}). The performance gap between the centralized approach and \algo may look counter-intuitive. However, similar results are reported by FeGAN~\cite{fegan}. The reason is that, \algo can see the data five times per epoch as compared to the centralized approach which only sees it once. This boosts the diversity of samples seen by \algo thereby providing superior performance.


We summarize the final statistical similarity and ML utility results of all three approaches and all four datasets in Tab.~\ref{table:5full_iid}. The scores are taken at the time in seconds when \central finishes 500 epochs training. One can see that \algo consistently achieves higher similarity (lower Avg-JSD and Avg-WD values) than the other two approaches. 
Similar results are obtained for ML utility.
F1-score difference shows that \algo outperforms (lower F1-score difference) all the baselines on all datasets, except on Credit where both \algo and \central achieve a perfect score.

\begin{table}[t]
\centering
\caption{Final similarity for \mdtgan and \algo and \central: 5 clients each having a complete data copy.}
\resizebox{\columnwidth}{!}{
\begin{tabular}{|C{1.4cm}|C{3cm}|C{3cm}| C{3cm}|}
\hline
\textbf{Dataset} & \textbf{Avg JSD \small(\mdtg/\alg/\central)} & \textbf{Avg WD \small(\mdtg/\alg/\central)} & \textbf{F1-score diff. \small(\mdtg/\alg/\central)}\\
\hline
{Adult}    & 0.072/\textbf{0.059}/0.117&  0.014/\textbf{0.012}/0.015  &0.101/\textbf{0.026}/0.056   \\
{Covertype} & 0.038/\textbf{0.018}/0.075 & 0.022/\textbf{0.021}/0.086 &0.493/\textbf{0.376}/0.460     \\
{Credit}   & 0.083/\textbf{0}/0.012 &\textbf{0.006}/\textbf{ 0.006}/0.041   &0.064/\textbf{0/0}  \\
{Intrusion}   & 0.095/\textbf{0.031}/0.032&  0.027/\textbf{0.02}/0.026  &0.085/\textbf{0.074}/0.075 \\
\hline
\end{tabular}
}
\label{table:5full_iid}
\vspace{-1em}
\end{table}

\subsubsection{Imbalanced amount of IID data}
\label{ssec:imbalanced}
For this experiment, we design a scenario where the number of data rows distributed among clients is highly imbalanced. Specifically, we include 5 clients in the group. 4 out of 5 clients contain only 500 rows of data randomly sampled from the original dataset. The last client contains the full dataset. We select 500 because it is the batch size setting in CTGAN. So we need at least 500 rows to form one mini batch for one epoch. 
This scenario is to show the effect of the model aggregation weights that the federator calculates during initialization. 

\begin{figure}[t]
	\begin{center}
		\subfloat[Avg-JSD by epoch]{
			\includegraphics[width=0.46\columnwidth]{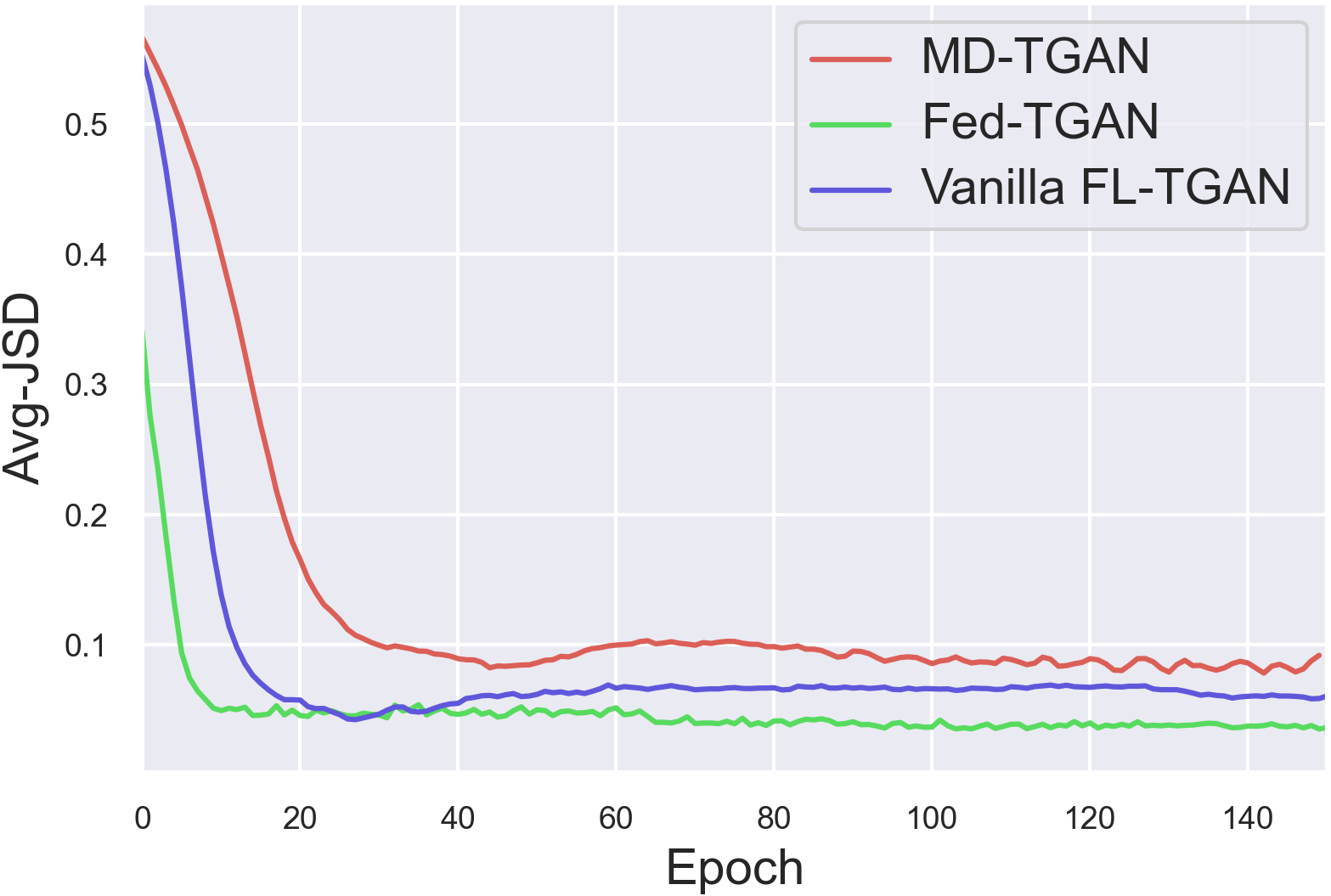}
			\label{fig:5imbalance_intrusionjsd}
		}
		\subfloat[Avg-JSD by time]{
			\includegraphics[width=0.47\columnwidth]{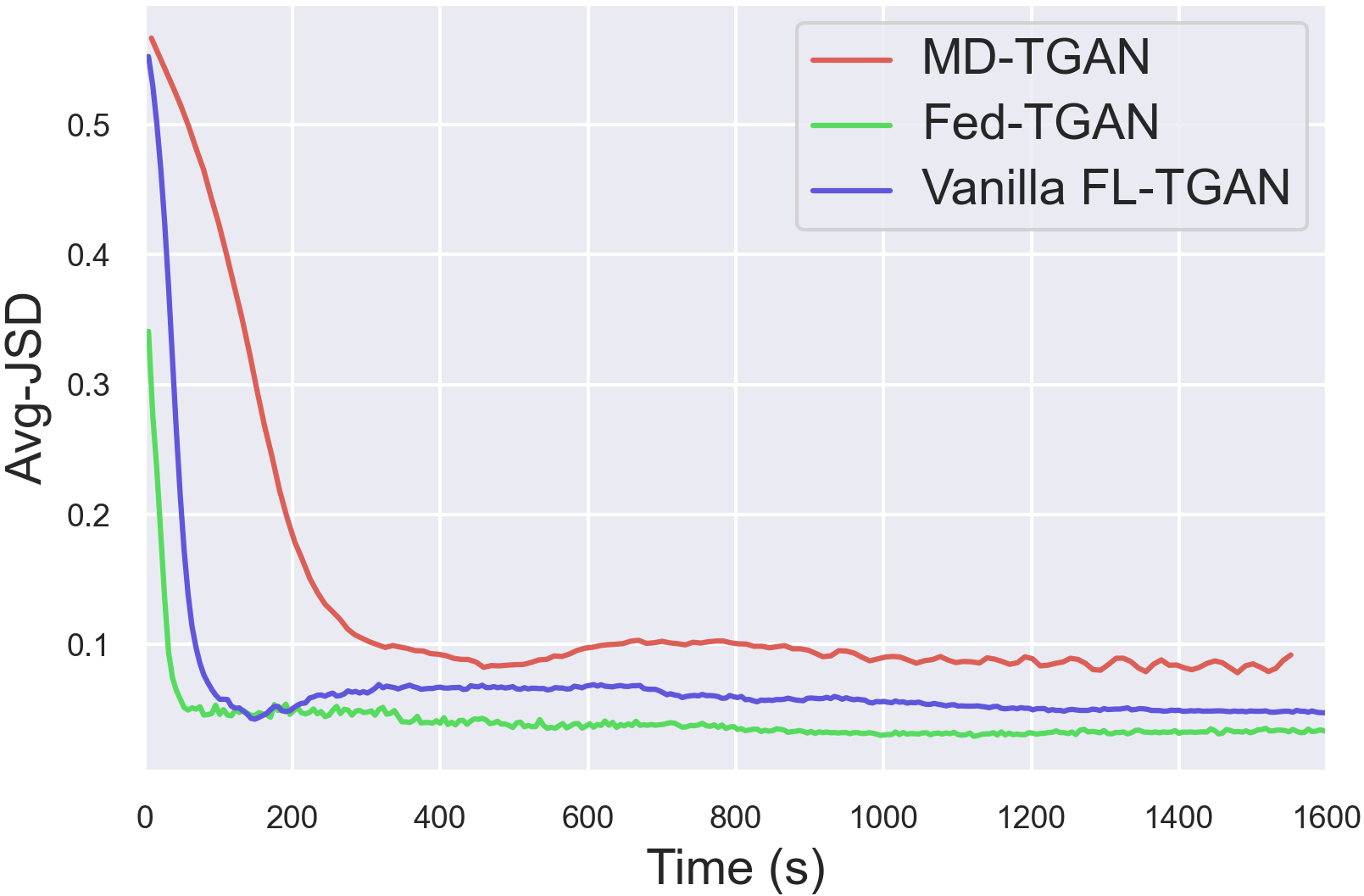}
			\label{fig:5imbalance_intrusionjsdt}
		}
		\\[-2ex]
	    \subfloat[Avg-WD by epoch]{
			\includegraphics[width=0.46\columnwidth]{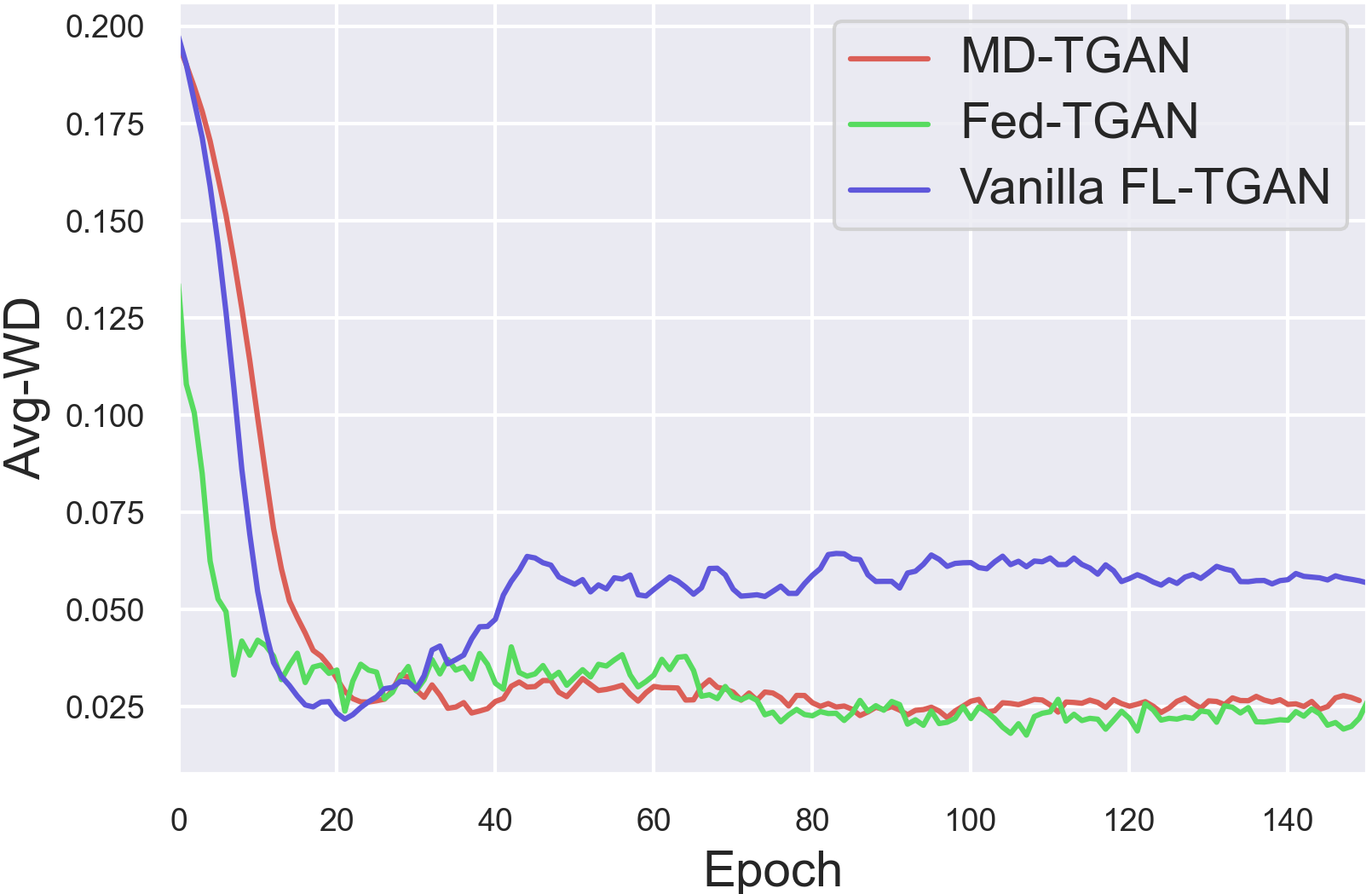}
			\label{fig:5imbalance_intrusionwd}
		}
		\subfloat[Avg-WD by time]{
			\includegraphics[width=0.47\columnwidth]{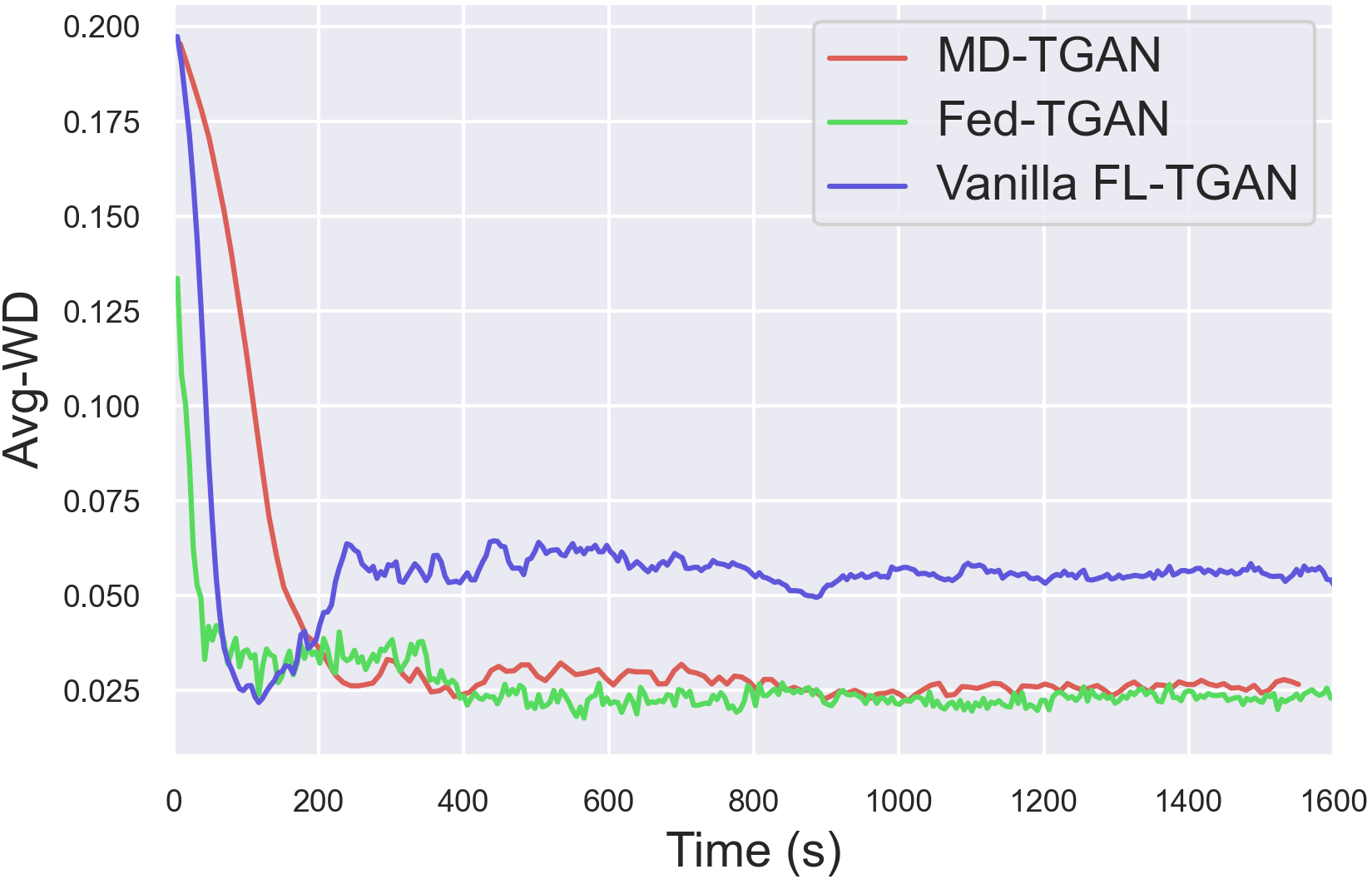}
			\label{fig:5imbalance_intrusionwdt}
		}
		\caption{MD-TGAN, \algo and vanilla FL-TGAN: 4 clients have 500, 1 client has 40k rows of sampled IID data.}
		\label{fig:5imbalance_intrusion}
	\end{center}
 	\vspace{-1.2em}
\end{figure}

Results are shown in Fig.~\ref{fig:5imbalance_intrusion}. For categorical columns, one can see that the Avg-JSD by epoch, converges faster for \algo than \vanilla by around 35\% (epoch 17 versus epoch 26) (see Fig.~\ref{fig:5imbalance_intrusionjsd}). 
Moreover, the Avg-JSD value for \algo after convergence is also smaller as compared to that of \mdtgan and \vanilla. Similar results are presented for measuring the Avg-JSD by time as well (see Fig.~\ref{fig:5imbalance_intrusionjsdt}). For continuous columns, 
\algo converges faster at very beginning.
Between 80 and 400 seconds, \algo is slightly worse than \mdtgan and \vanilla. From then on till the end, \mdtgan and \algo perform similarly, see Fig.~\ref{fig:5imbalance_intrusionwdt}. A similar pattern can be found also while computing the Avg-WD with respect to epochs, see Fig.~\ref{fig:5imbalance_intrusionwd}.

Full results on the four datasets are presented in Tab.~\ref{table:5imbalance_iid}. The ML utility results are also inline with previous case with only slightly degraded scores which reflect the added difficulty of this scenarios. Regarding data similarity we notice that except for the Intrusion dataset, \algo and \vanilla perform similarly for continuous columns. 
However for categorical columns, \algo outperforms \vanilla for most datasets. 
\algo converges better than \vanilla because the model trained on 40K converges better than model trained on 500, since all their data are IID and sampled from original dataset. As we give more weight to the model which is better trained, we benefits from its better convergence. 
The reason that \algo and \vanilla have similar performance on Adult dataset can be due to the fact that Adult dataset has less columns, thus is simpler to learn. The Avg-WD results for Adult, Covertype and Credit datasets are similar, the reason is because in each of the 500 IID data, continuous columns distributions are well maintained as in original data.  
From Fig.~\ref{fig:5imbalance_intrusion} and Tab.~\ref{table:5imbalance_iid} (results are taken at the time when MD-TGAN finishes 150 epochs training), we conclude that under an imbalanced data quantity distribution across clients, \vanilla not only suffers from slow convergence, but also results in poor sample quality. 


\begin{table}[t]
\centering
\caption{Final similarity for \mdtgan, \algo and \vanilla: 4 clients have 500, 1 client has 40k rows of sampled IID data.}
\vspace{-0.5em}
\resizebox{\columnwidth}{!}{
\begin{tabular}{|C{1.4cm}|C{3cm}|C{3cm}|C{3cm}|}
\hline
\textbf{Dataset} & \textbf{Avg JSD \small(MD/Fed/Vanilla-FL)} & \textbf{Avg WD \small(MD/Fed/Vanilla-FL)}  & \textbf{F1-score diff. \small(MD/Fed/Vanilla-FL)}\\
\hline
{Adult}    &0.07/\textbf{0.062/0.062} &0.014/\textbf{0.012/0.012 } &0.130/\textbf{0.032}/0.036  \\
{Covertype} &0.029/\textbf{0.026}/0.032 &\textbf{0.02/0.02/0.02} &0.503/\textbf{0.417}/0.419    \\
{Credit}   &0.078/\textbf{0.007}/0.011&0.006/\textbf{0.005/0.005} &0.132/\textbf{0/0} \\
{Intrusion}  &0.092/\textbf{0.037}/0.044 &\textbf{0.025}/\textbf{0.025}/0.052 &0.114/\textbf{0.092}/0.134\\
\hline
\end{tabular}
}
\label{table:5imbalance_iid}
\vspace{-1em}
\end{table}


\subsubsection{Ablation analysis}
\label{ssec:ablation}

Recall the weights calculation process in Fig.~\ref{fig:algo_weights}. The $SD_i$ is composed of two parts: (1) the ratio of the number of data rows locally available at the client $i$ to global number of data rows, i.e., $\frac{N_i}{N_{all}}$; and (2) the similarity calculated between the local data distribution of client $i$ and the global distribution, i.e., $1 - \frac{SS_i}{\sum_{i=1}^{P} SS_i}$. Our experiment in Sec.~\ref{ssec:imbalanced} shows the difference between \algo and \vanilla (i.e., \algo with equal weights for all clients). Results show that weighting clients differently based on the amount of data is indeed useful when the data quantity at each client is skewed.
The contribution of data number ratio part is intuitive. Therefore in this ablation analysis,  we design a scenario where for \algo, the client weights are only calculated using data number ratio of each client, without using the similarity component.

\begin{figure}[t]
\vspace{-0.5em}
	\begin{center}
		\subfloat[Avg-JSD by epoch]{
			\includegraphics[width=0.47\columnwidth]{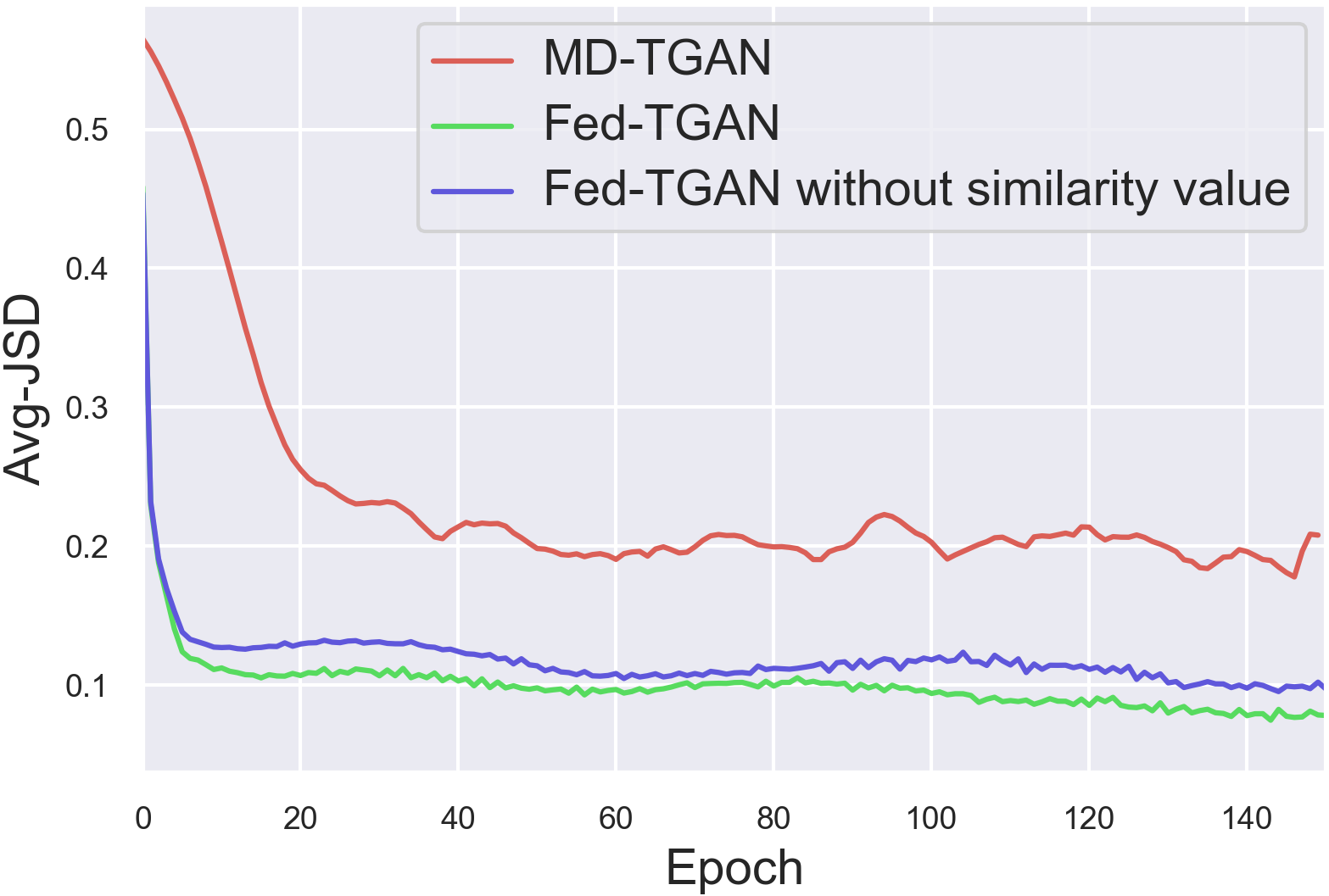}
		}
		\subfloat[Avg-JSD by time]{
			\includegraphics[width=0.47\columnwidth]{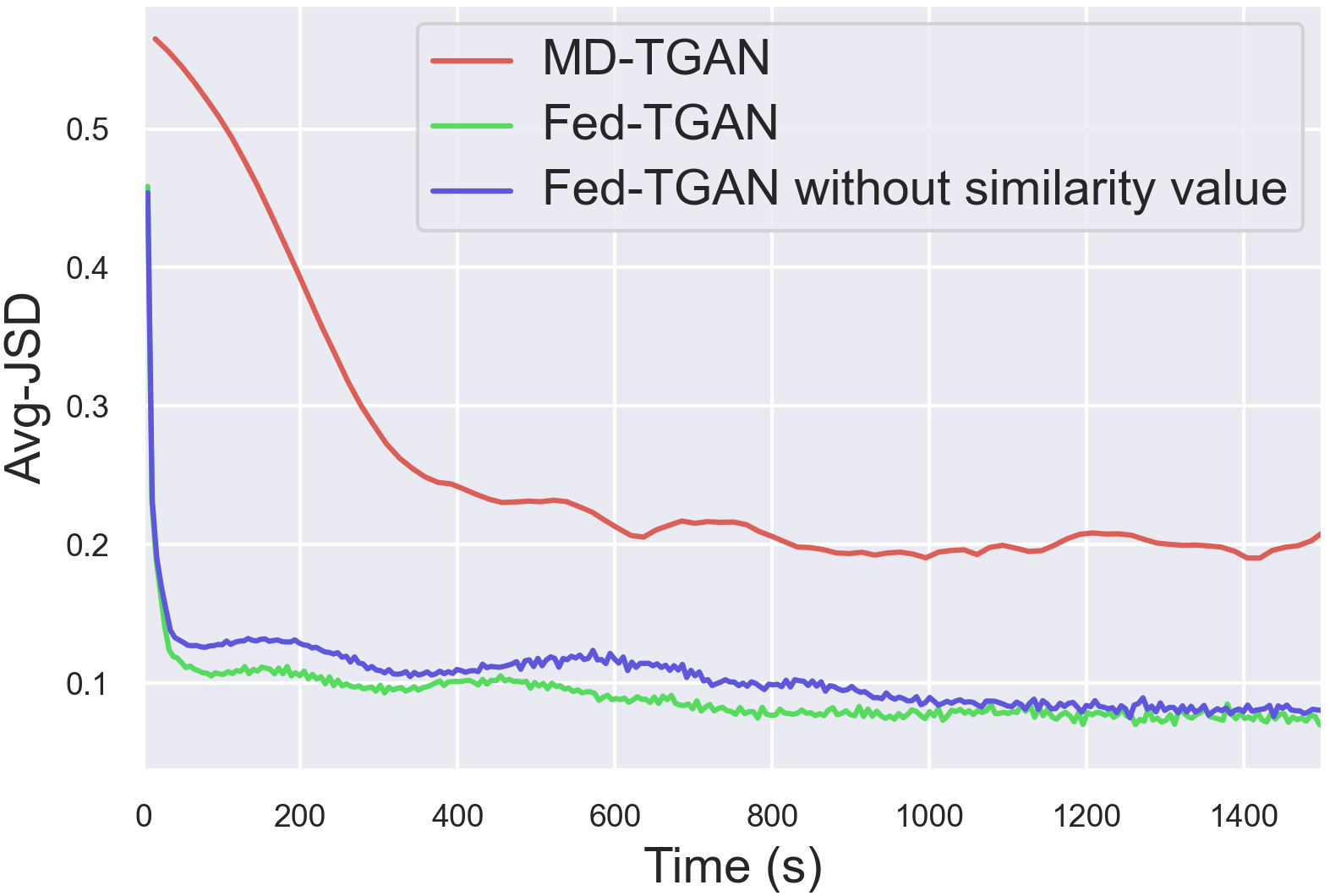}
		}\\[-1em]
				\subfloat[Avg-WD by epoch]{
			\includegraphics[width=0.47\columnwidth]{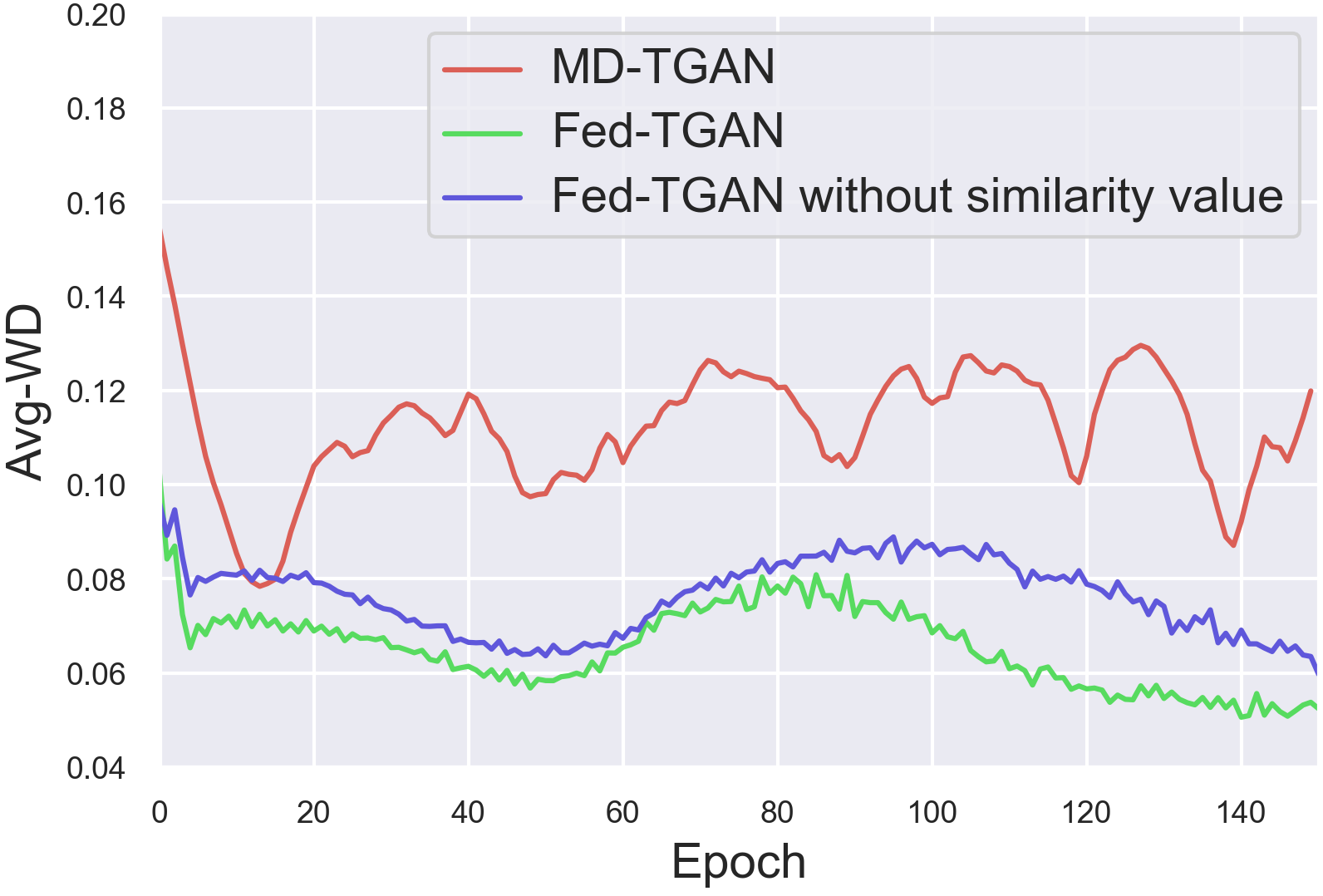}
			\label{fig:ablation_c}
		}
		\subfloat[Avg-WD by time]{
			\includegraphics[width=0.47\columnwidth]{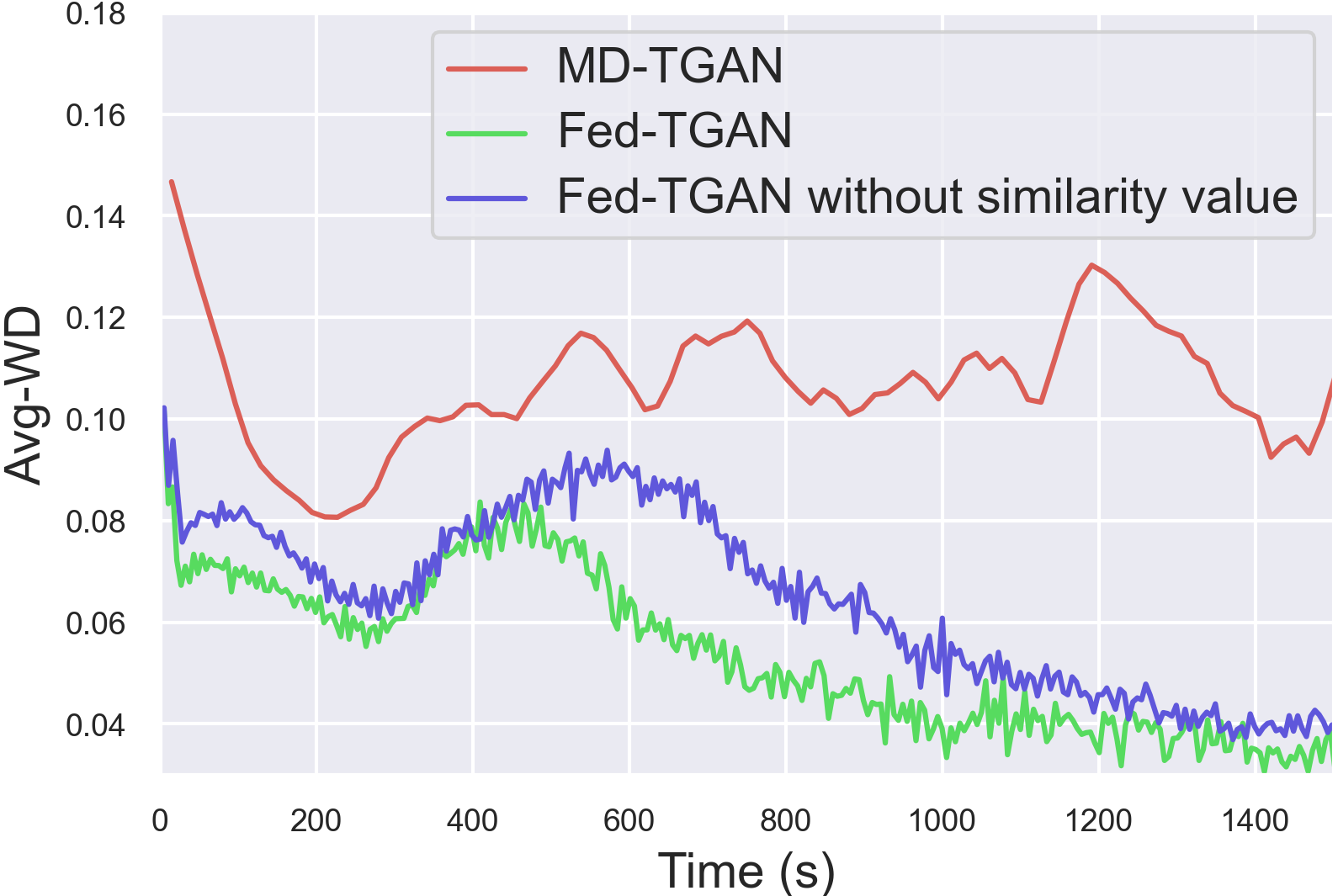}
			\label{fig:ablation_d}
		}
		\caption{\mdtgan, \algo and \algo without similarity weights: 4 clients have 10k, 1 client has 40K rows sampled \textit{Non IID} data.}
		\label{fig:5ablation_intrusion}
	\end{center}
  	\vspace{-0.5em}
\end{figure}

\begin{table}[t]
\centering
\caption{Final similarity for \mdtgan,  \algo and \algo without similarity weights (Fed$\backslash$SW): 4 clients have 10k, 1 client has 40K rows of sampled \textit{Non IID} data.}
\resizebox{\columnwidth}{!}{
\begin{tabular}{|C{1.4cm}|C{3cm}|C{3cm}|C{3cm}|}
\hline
\textbf{Dataset} & \textbf{Avg JSD \small(MD/Fed/Fed$\backslash$SW)} & \textbf{Avg WD \small(MD/Fed/Fed$\backslash$SW)}  & \textbf{F1-score diff. \small(MD/Fed/Fed$\backslash$SW)} \\
\hline
{Adult}    & 0.37/\textbf{0.149}/0.261&0.107/\textbf{0.026}/0.027 &0.179/\textbf{0.065}/0.080   \\
{Covertype} & 0.089/\textbf{0.05}/0.06 &0.125/\textbf{0.045}/0.056 &0.747/\textbf{0.466}/0.468   \\
{Credit}   &0.074/\textbf{0.014}/0.06&0.04/\textbf{0.01}/0.015  &0.154/\textbf{0.121}/0.138\\
{Intrusion}  &0.208/\textbf{0.068}/0.073 &0.107/\textbf{0.032}/0.036 &0.584/\textbf{0.095}/0.117\\
\hline
\end{tabular}
}
\label{table:5ablation}
\vspace{-0.5em}
\end{table}

To better show the importance of similarity weights, we design a specific scenario for this experiment. Still with 5 clients, 4 of them containing 10k IID data sampled from original data, the last client is modified to contain 40k rows of data by repeating one row sampled from the original dataset 40k times. One can imagine, this last client has a large number of rows, but contains little information in them. Fig.~\ref{fig:5ablation_intrusion} shows the results on Intrusion dataset. 
One can already notice that this scenario badly hits \mdtgan since it treats all clients equally while updating the generator's weights. Moreover, for the results in Fig.~\ref{fig:ablation_c} and \ref{fig:ablation_d}, one can see the client with 40k repeated data introduces oscillation to the curves of \algo with and without the similarity component. As expected, the curve for \algo without similarity component naturally performs worse than \algo. Results in Tab.~\ref{table:5ablation} (Scores are taken at the time when MD-TGAN finishes 150 epochs training.) shows that \algo undoubtedly outperforms \mdtgan and \algo without similarity computation for all datasets. Therefore, similarity component in \algo gives more stability for model convergence. 








\subsection{Training time analysis}
\label{ssec:training_time}
\begin{figure}[t]
    \vspace{-0.7em}
	\begin{center}
		\subfloat[Training time per epoch]{
			\includegraphics[width=0.49\columnwidth]{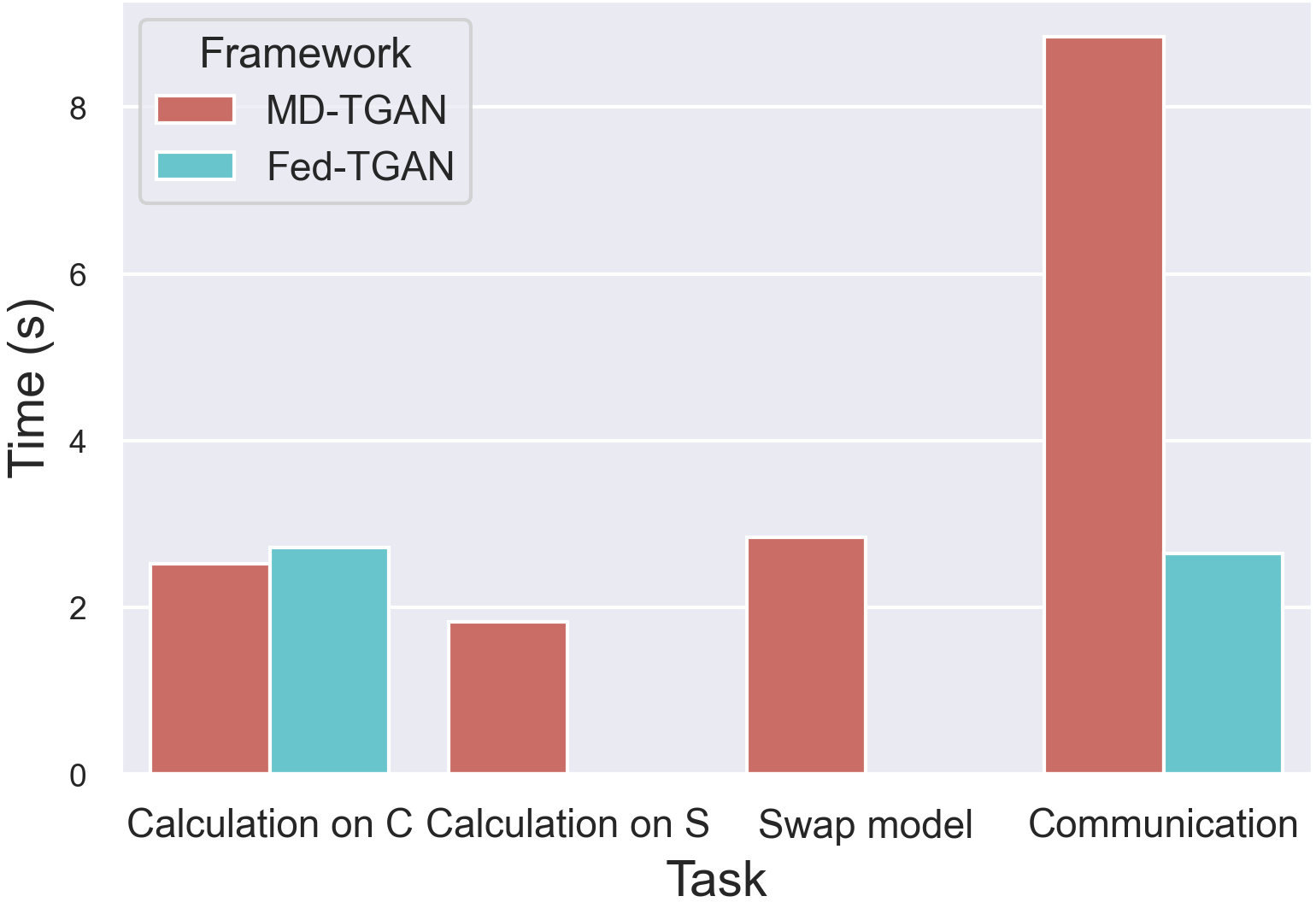}
			\label{fig:time_per_epoch}
		}
		\subfloat[Impact of epochs per round]{
			\includegraphics[width=0.5\columnwidth]{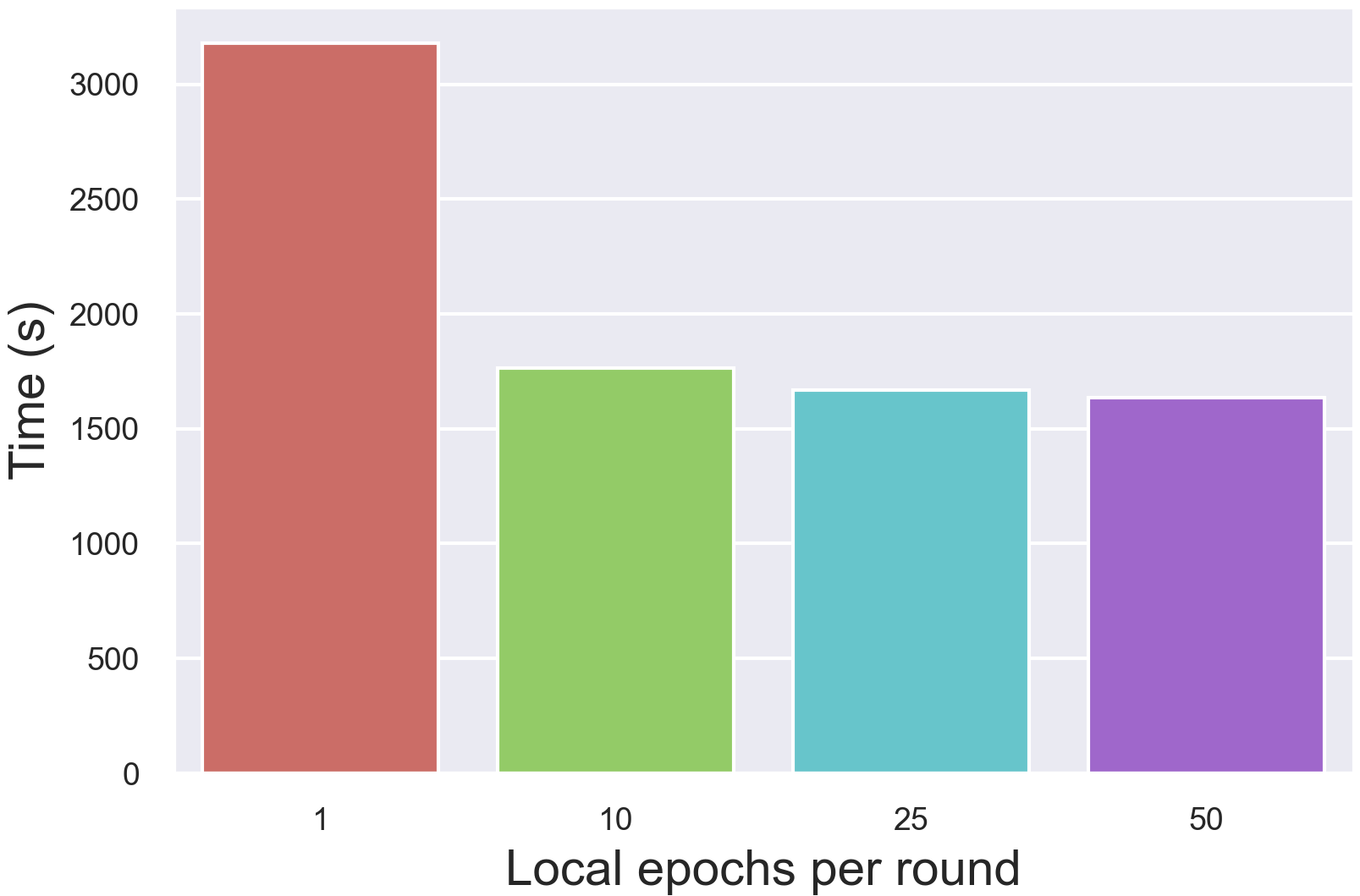}
			\label{fig:varying_epochs}
		}
		\vspace{-0.7em}
		\caption{(a) Epoch training time per phase of \mdtgan and \algo with 5 clients. \textit{S} for server, \textit{C} for client. (b) Total training time with varying local epochs per round for \algo.}
		\label{fig:timedistribution_intrusion}
	\end{center}
	\vspace{-1em}
\end{figure}

Above experiments all focus on the quality of generation. In this section, we study the training efficiency of MD-TGAN and \algo. The first experiment setup is the same as in Sec.~\ref{ssec:iid}. 
The entire FL system consists of 5 clients, and each of them possesses the full original dataset.  Fig.~\ref{fig:time_per_epoch} shows the time distribution for MD-TGAN and \algo during one training epoch on Intrusion dataset. The \textit{Calculation on C} is calculated from the start of training of first client to the finishing of all clients.
For \algo, after aggregating the local models and sending back the global model, all clients can immediately start to train for the next round. But for MD-TGAN, all clients' training need to wait for synthesized data from generator. That adds times for its \textit{Communication}.  
\textit{Communication} counts the time for exchanging model weights, swapping discriminator between clients, or sharing training data between server (or federator) and clients. 
We can see that for \algo, the calculation time on federator is negligible since it is only averaging model weights. \algo has a slightly higher calculation time on clients because it trains both generator and discriminator networks on clients. 
The communication time of MD-TGAN is much higher, because for updating generator or discriminator in MD-TGAN, the generator needs to send the generated data from generator to each discriminator. Since MD-TGAN only has one server, the above tasks can not be distributed. Fig.~\ref{fig:time_per_epoch} shows that the communication time of \algo is only 30\% of what MD-TGAN uses. \algo saves more than 200\% of the time taken by MD-TGAN per epoch.






\begin{figure}[t]
    \vspace{-1em}
	\begin{center}
		\subfloat[Avg-JSD by time]{
			\includegraphics[width=0.47\columnwidth]{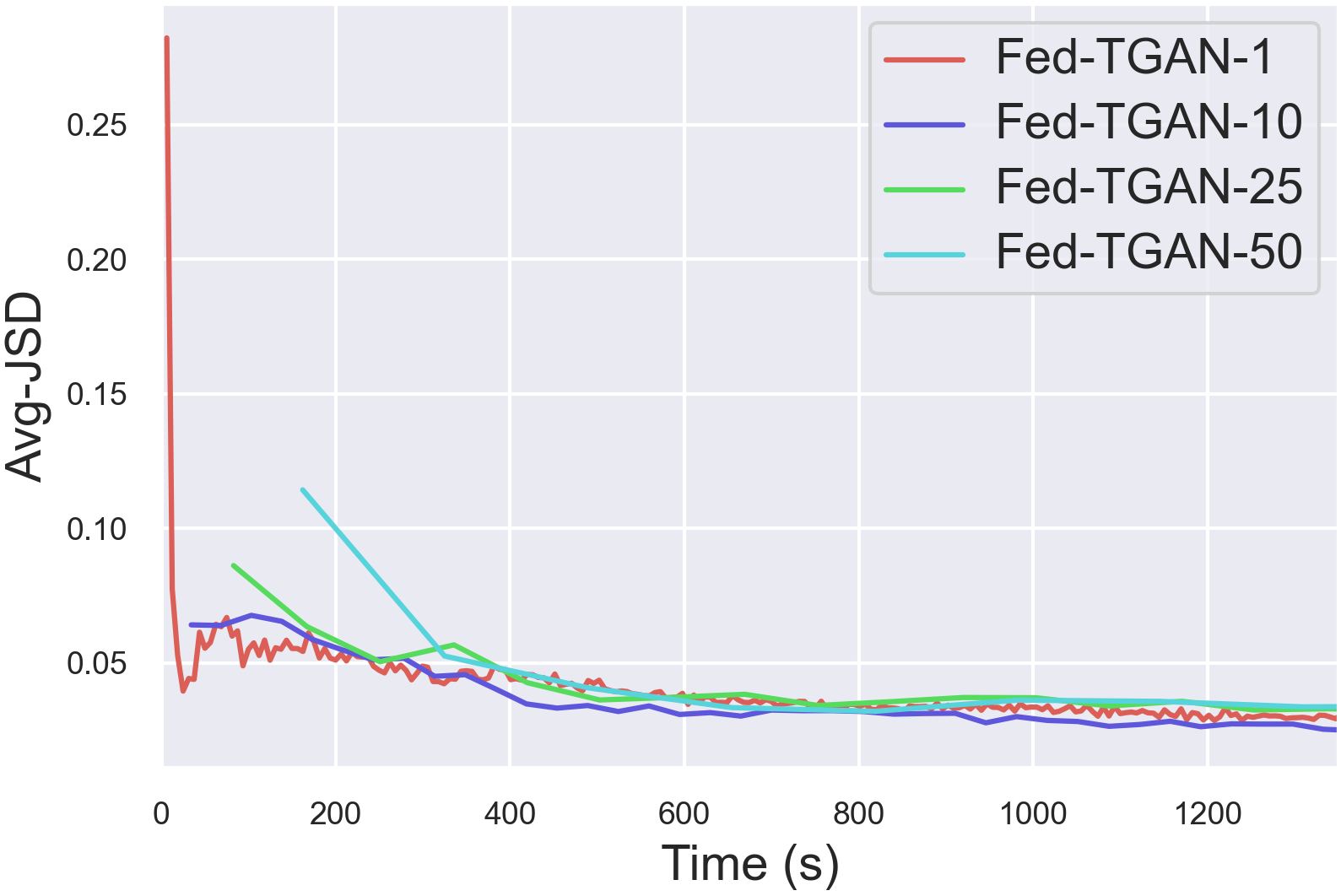}
			\label{fig:time_analysis_variation_interval_jsd}
		}
		\subfloat[Avg-WD by time]{
			\includegraphics[width=0.47\columnwidth]{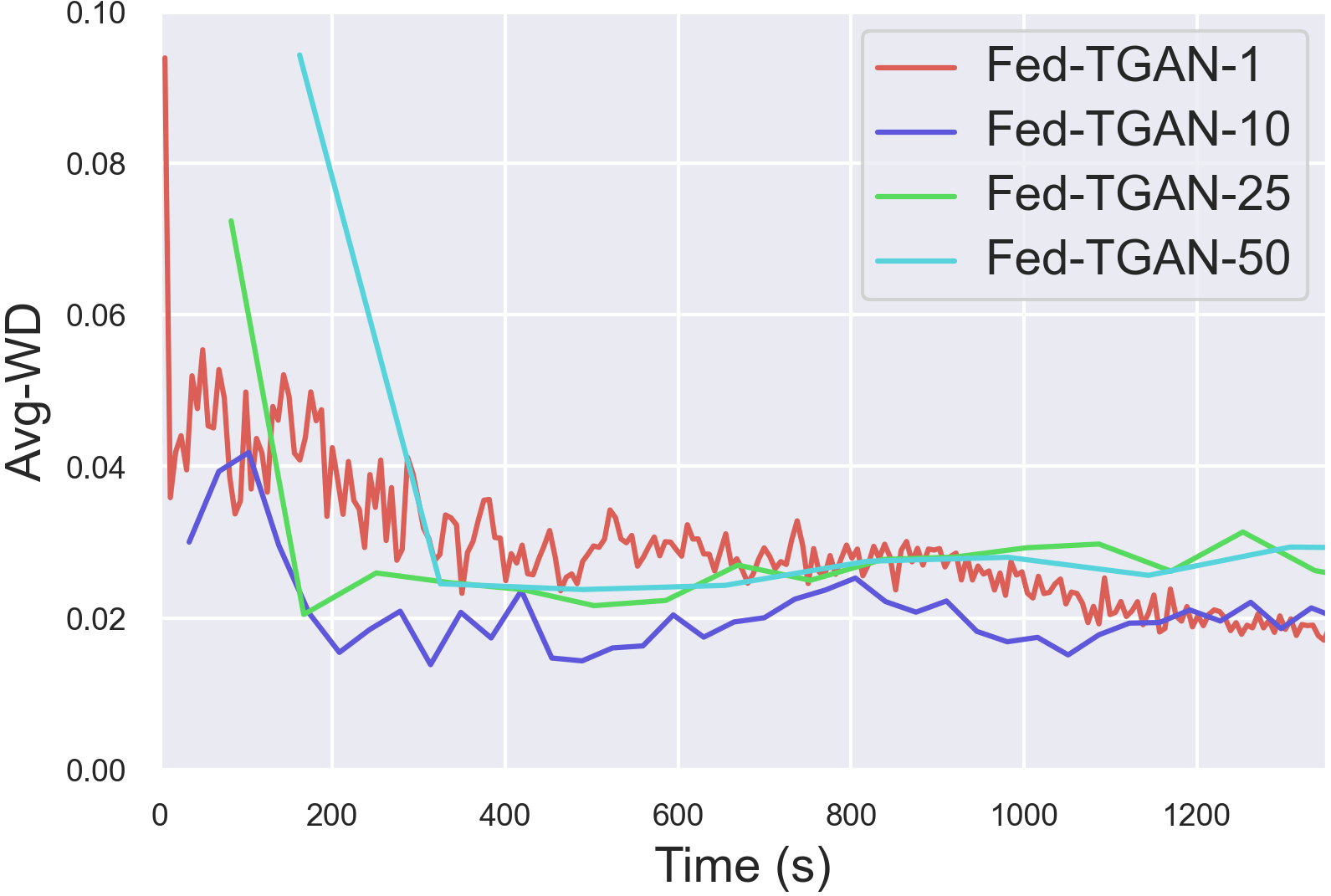}
			\label{fig:time_analysis_variation_interval_wd}
		}
		\vspace{-0.7em}
		\caption{Varying epochs per round with 5 clients.}
		\label{fig:timedistribution_interval}
	\end{center}
	\vspace{-1.2em}
\end{figure}

In the second experiment, instead of aggregating models at the end of each epoch for every client, we vary the number of local training epochs before aggregating models for \algo. 
Fig.~\ref{fig:varying_epochs} shows the total training time with variation on local epochs per round for \algo, where we fix all the clients to train for 500 epochs in total. Therefore with more local epochs per round, we are left with less rounds for the whole training process. For local epochs 1, 10, 25 and 50, the corresponding round numbers are 500, 50, 20 and 10. 
The massive time is decreased between local epoch 1 and others is simply because of the reduction of model aggregations. The differences among other numbers of local epochs are not that significant. 
Fig.~\ref{fig:timedistribution_interval} shows the generation results under different local epochs. We see for categorical columns, the Avg-JSD converges for all to a small value. For continuous columns, the Avg-WD for \algo with 10 local epochs per round converges fastest and provides the best result until 1150s. This result indicates that it is possible to speed up training of \algo by utilizing more local epochs while still preserving the statistical similarity between real and synthetic datasets. However, increasing the local epochs to a large value can potentially lead to over-fit on the local data of clients ultimately deteriorating performance. Thus, the local epoch number introduces a trade-off between efficiency and performance.



\begin{figure}[t]
    \vspace{-0.5em}
	\begin{center}
		\subfloat[Scalability on number of clients]{
			\includegraphics[width=0.47\columnwidth]{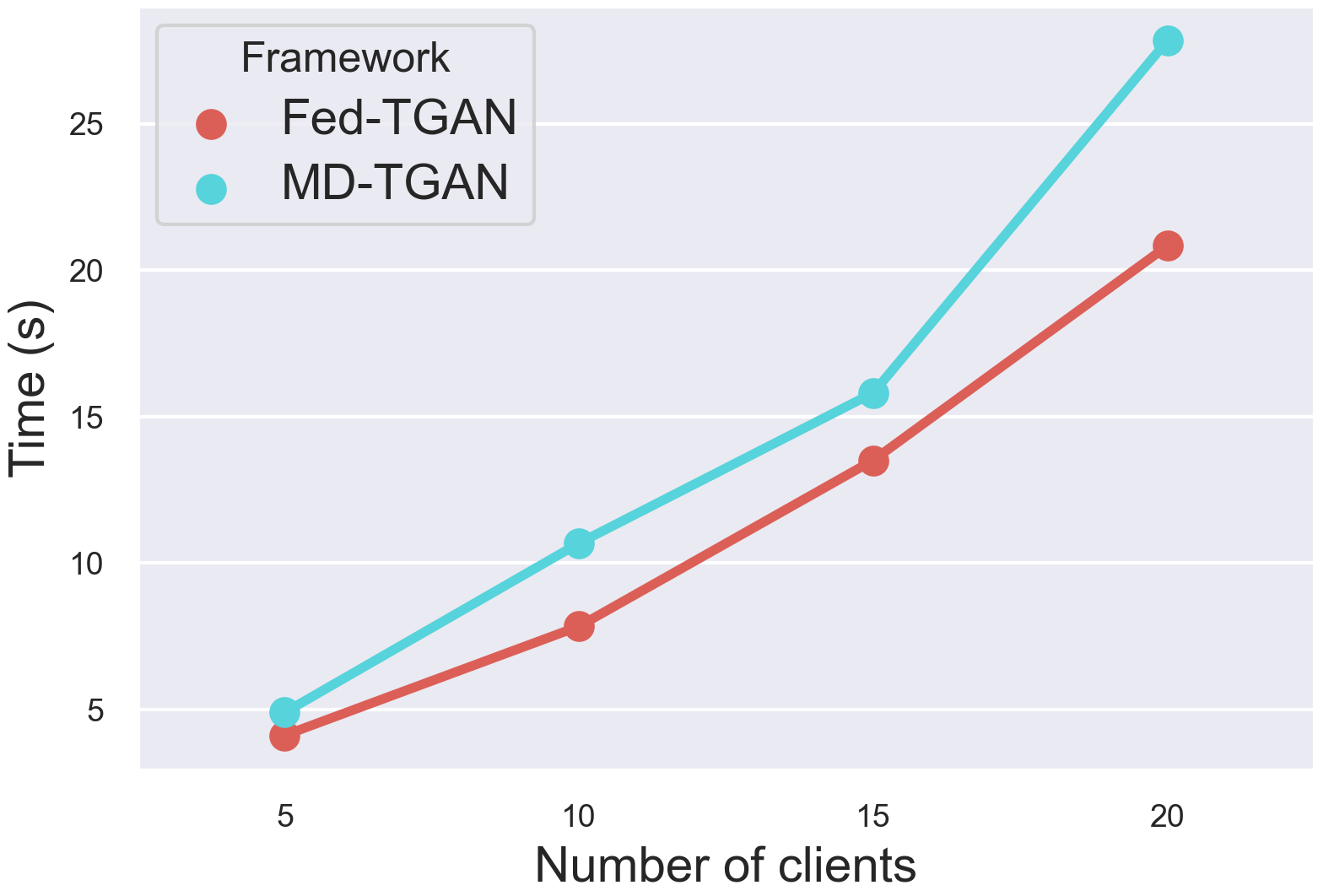}
			\label{fig:varying_number}
		}
		\subfloat[Scalability on data per client]{
			\includegraphics[width=0.5\columnwidth]{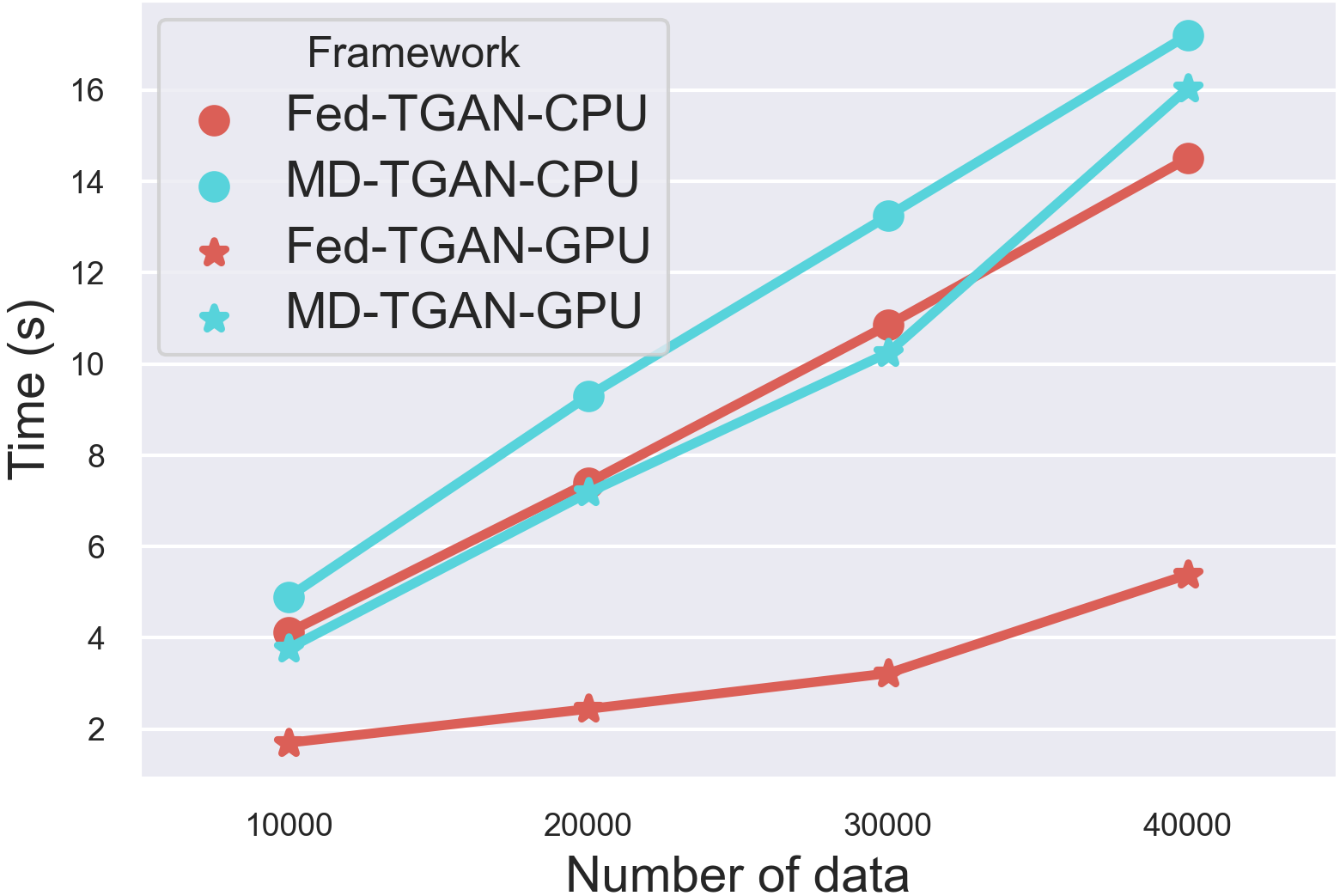}
			\label{fig:varying_data}
		}
		\caption{\mdtgan and \algo with: (a) Varying number of clients and fixed data per client. (b) Fixed 5 clients and varying amounts of data per client.}
		\label{fig:time_intrusion}
	\end{center}
	\vspace{-1em}
\end{figure}

Next, we evaluate time consumption of one epoch for \algo and MD-TGAN with varying factors. First, we fix the number and type of data (10K IID data from Intrusion dataset) on each client, and vary the number of clients from 5 to 20. For computing resource limitations, all the experiments with varying number of clients are implemented using CPUs on client side with the server (or federator) side using the GPU. To limit interference between different processes, CPU affinity is used to bind each client to one logical CPU core. 
Fig.~\ref{fig:varying_number} clearly shows that \algo scales better than \mdtgan with number of clients. In \mdtgan the central server becomes increasingly the bottleneck when adding clients due to large amount of data exchanged with each client. Second, we fix the number of clients to 5 and vary the number of IID sampled data from the Intrusion dataset. We vary the number of rows from 10k to 40k. The experiments are implemented on both CPU and GPU for clients. The result in Fig.~\ref{fig:varying_data} shows that with increasing the number of data rows on each client, \algo and \mdtgan both experience an increase in the training time per epoch. The difference between the two algorithms is small when training happens on CPU. But when using GPU, we have an increasing difference with increasing amount of data on clients. The reason is because when sharing data between the client and the server, tensors that are on GPU must first be detached from GPU to CPU for being sent through the Pytorch RPC framework. Since \algo trains all tabular GAN models locally on each client, the training process is highly accelerated by GPUs. As clients in the federated setting only need to detach the model from the GPU to CPU at the end of training to share them with the when exchanging messages. And so, since the server shares message more times in MD-TGAN than the federator in \algo, the GPUs do not accelerate the training process of MD-TGAN as much as for \algo.


\section{Conclusion}
Due to ever increasing distributed data sources and privacy concerns, it is imperative to learn GANs in a decentralized and privacy-preserving manner -- features offered by federated learning systems. While the prior art demonstrates the feasibility of learning image GANs in FL systems, it remained unknown if the predominant tabular data in industry and its GANs can be deployed in a FL framework. This paper proposes and implements, \algo, a first of its kind FL architecture and prototype for tabular GANs, overcoming specific challenges related to tabular data. Two main features of \algo are (i) privacy preserving feature encoding to enable model initialization across heterogeneous data sources, and (ii) table-similarity aware weighting for merging local models. We extensively evaluate \algo using a state-of-the-art tabular GAN and compare it with two alternative decentralized architectures, i.e., \mdtgan and \vanilla, and a centralized approach. Our results show that \algo can generate synthetic tabular data that preserves high similarity to the original data with faster convergence speeds, even in the challenging case of Non-IID data among clients. The prototype of \algo is currently under testing by a fortune 500 financial institute. The promising evaluation results confirm that \algo can help large organizations to unlock their data stored across multi-national silos to build a better tabular data synthesizer in a privacy preserving manner. We plan to release the source code after publication of the paper.


\ifnotdoubleblind

\fi

\bibliographystyle{abbrv}
\bibliography{sigproc}  
\ifnotdoubleblind
\fi
\clearpage
\pagebreak
\appendix
\section{Reproducibility of the Fed-TGAN system}
\subsection{Target audiences and potential impact}
Privacy regulations such as the GDPR are implemented to protect personal data by limiting the sharing of data. This hinders activities to discover valuable knowledge from the data. Synthetic data is a promising solution to this conundrum which is rapidly gaining traction among multi-national corporations and startups alike. Data analysts, such as engineers and data scientists working in fields such as e-commerce, banking and insurance, will be interested in this topic. 
A report\footnote{\url{https://elise-deux.medium.com/the-list-of-synthetic-data-companies-2021-5aa246265b42}} shows that during 2016 to 2020, the synthetic data market has attracted more than 210 million dollars. 
It is forcasted~\footnote{\url{https://blogs.gartner.com/andrew_white/2021/07/24/by-2024-60-of-the-data-used-for-the-development-of-ai-and-analytics-projects-will-be-synthetically-generated/}} that by 2024, 60\% of the data used for the development of AI and analytics projects will be synthetically generated.
Data scientists in such synthetic data companies will also be interested in this work.
The Fed-TGAN framework proposed in this paper is a part of services of 
\textbf{Generatrix}\footnote{\url{https://www.generatrix.ai}} startup, 
which is currently collaborating with world-class e-commerce companies, banks and asset management companies to boost their data sharing within and outside their companies and reduce the burden of data protection officers (DPO) of all the companies.


\subsection{Reproducibility}
The project is temporarily hosted on google drive \footnote{\url{https://drive.google.com/file/d/1ggjCrywVwk-8MG5Oq4TpYzohyqgz7cm3/view?usp=sharing}}. We will publish the code on github after acceptance of this paper. 

\subsubsection{Prerequisites}
Experiments are run under Ubuntu 20.04 on two machines. Each machine is equipped with 32 GB memory, GeForce RTX 2080 Ti GPU and 10-core Intel i9 CPU. Each  CPU core has two threads, hence each machine contains 20 logical CPU cores in total. The machine are interconnected via 1G Ethernet links (measured speed: 943Mb/s). The main required libraries and corresponding versions are as follows:
\begin{align*}
    python==3.7.10\\
    torch==1.9.1\\
    numpy==1.21.0\\
    pandas==1.2.4\\
    sklearn==0.24.1\\
    dython=0.6.4\\
    scipy==1.4.1\\
\end{align*}
The code is also tested under macOS Mojave and python3.7.9. GPU is not mandatory for the experiments, but recommended for faster training times.

\subsubsection{Implementation}
The project file contains two folders: \textbf{Server} and \textbf{Client}. The \textbf{Server} is the federator in federated learning (FL). The \textbf{Client} contains the FL participants. The testbed is a distributed framework realised using Pytorch RPC. The server and client(s) can be placed on different computers, as long as they can communicate with each other via a network. We first introduce how to train the FED-TGAN algorithm (using the \textit{Intrusion} dataset as example). Then explain how to evaluate the results (i.e., statistical similarity and ML utility) in the same way as in the paper.

{\bf Training}:
In this example we use the \textit{Intrusion} dataset together with 2 clients  running on the same host. 
To begin training with FedTGAN, we need to first start the aggregator. This is done entering the \textbf{Server}
folder and typing the following command:
\begin{lstlisting}[language=bash]
  $ python -m dtds.distributed -rank 0 -ip X.X.X.X -epoch 500 -world_size 3 -datapath "data/raw/Intrusion_train.csv"
\end{lstlisting}
where \textit{world\_size} represents the total number of participants (including the server) in the FL training group. In our case 3 represents 1 server plus 2 clients. \textit{rank} indicates the rank of each participant in the FL training group. The rank must be unique for each participant and the server must always be rank 0. 
$\mbox{X.X.X.X}$ is the \textit{ip} on which the server listens for connections from the clients.
\textit{epoch} indicates the number of global rounds  
to train the model. The \textit{epoch} in this setting equals to the notion FL \textit{round} in the paper. One epoch in this setting means federator aggregates one time of the client models.
Finally, the \textit{datapath} indicates the path to the training dataset for \textbf{clients side}. The data path \textbf{data/raw/Intrusion\_train.csv} specified in this command does not mean there is dataset at this path in server side. On the contrary, there should not exist dataset in server side. We specify the data path for client from server side is due to the mechanism of Remote Procedure Call (RPC). In Pytorch RPC, all the instances of clients are initialized in server side and then execute the function in client side. For the sake of simplicity, we directly initialize all the \textit{datapath} for all the clients. Because of that, when starting the client process, all the clients should contain dataset in the relative path \textbf{data/raw/Intrusion\_train.csv}. The content of dataset \textbf{Intrusion\_train.csv} under different client can be different, but the name of the dataset should be the same.


{\bf Training in Client side:} Under \textbf{Client} folder, we already provided two client folders: \textbf{Client0} and \textbf{Client1}. \textbf{Client0} and \textbf{Client1} can be run on two different machines, and none of them is  necessarily located in the same machine as the server, our demo works as long as there is network connection between the server and clients. 
Actually, there is no difference between the two folders. If anyone wants to add more clients to the FL system, just copy and paste this folder to your local machine. For clients to join the FL, under both \textbf{Client0} and \textbf{Client1} folders, run:
\begin{lstlisting}[language=bash]
  $ python -m dtds.distributed -ip X.X.X.X -rank 1 -world_size 3
\end{lstlisting}
and 
\begin{lstlisting}[language=bash]
  $ python -m dtds.distributed -ip X.X.X.X -rank 2 -world_size 3
\end{lstlisting}
the \textit{ip} is the ip address of server. We can observe that for two clients, the only difference is that clients need to self-claim a unique rank for themselves. And for $\textbf{N}$ clients, their ranks should start from $1$ to $\textbf{N}$ without repetitions.

{\bf Troubleshooting}: One possible error is 
\begin{lstlisting}[language=bash]
  $ RuntimeError: ECONNREFUSED: connection refused
\end{lstlisting}
We use Gloo as backend. By default, Gloo will try to find the right network interface to use, but sometimes it is not correct. In that case, the network interface can be overridden using the following environmental variable:
\begin{lstlisting}[language=bash]
  $ export GLOO_SOCKET_IFNAME=eth0
\end{lstlisting}
where \textbf{eth0} is the correct network interface name.

\subsubsection{Evaluation of the result}
We still use \textit{Intrusion} dataset as example.  It is worth to notice that in the following text. The data path \textbf{Server\textbackslash Intrusion\_result\textbackslash} and synthetic dataset names are manually defined in the \textit{sample\_data()} function within the script \textbf{Server\textbackslash dtds\textbackslash distributed.py}. During the training, under the folder \textbf{Server\textbackslash Intrusion\_result\textbackslash},
it will save the generated synthetic data at the end of each training round. For instance, if we set \textit{epoch} to 500, then there will have datasets named from \textbf{Intrusion\_synthesis\_epoch\_0.csv} to \textbf{Intrusion\_synthesis\_epoch\_499.csv} under \textbf{Server\textbackslash Intrusion\_result\textbackslash} folder. These files are the synthesized data from global model at the end of each training epoch.  All the synthesized data is the same size of the real data because they will be used to do the evaluation (i.e., compare with real data) later. A file named \textbf{timestamp\_experiment.csv} will be generated after successfully finishing the training process under \textbf{Server} folder. This file will contain one column with 500 rows, each value represents the total training time of the corresponding training epoch.

{\bf Statistical similarity:} For statistical similarity, going to \textbf{Server} folder and run: 
\begin{lstlisting}[language=bash]
  $ python similarity_analysis.py -nepoch 500
\end{lstlisting}
\textit{nepoch} represents the number of training epochs. This parameter is needed to indicate how many synthetic dataset generated during the training. Above command line will generate a file \textbf{Intrusion\_statistical\_similarity\_analysis.csv} under \textbf{Server} folder. The content is the same format as in Tab.~\ref{tab:appendix_tab1}. Each row shows the statistical similarity between  \textbf{Intrusion\_synthesis\_epoch\_\textit{E}.csv} and real data where \textbf{\textit{E}} represents \textit{Epoch\_No.}. \textit{time\_stamp} represents the accumulated training time at the end of \textit{Epoch\_No.} epoch. The \textbf{Avg\_JSD} and \textbf{Avg\_WD} are the two metrics we used for example in Tab.~\ref{table:5full_iid}.

{\bf Machine Learning (ML) utility:}  For ML utility, going to \textbf{Server} folder and run:
\begin{lstlisting}[language=bash]
$ python utility_analysis.py -train_path 'data/raw/Intrusion_train.csv' -test_path 'data/raw/Intrusion_test.csv' -synthetic_path 'Intrusion_result/Intrusion_synthesis_X.csv'
\end{lstlisting}
\textit{train\_path} indicates the data path to real data that is used to train FL model. \textit{test\_path} also indicates the data path to part of real data, this part is not used for training.  \textit{synthetic\_path} introduces the path to synthetic data, user needs to change \textit{X} to the corresponding synthetic data. In general, user does not need to change the path for \textit{train\_path} and \textit{test\_path} for this demo. But for \textit{synthetic\_path}, user should check the existence of the synthetic data. After running above command line, user should see the output from terminal similarly as follows:
In the terminal, after evaluation, you will see the output like:
\begin{lstlisting}[language=bash]
$ difference in f1-score: 0.0849535625139106
\end{lstlisting}
This metric is the \textbf{F1-score diff.} we used for example in Tab.~\ref{table:5full_iid}.
 

\begin{table}
\centering
\caption{Statistical similarity result format.}
\begin{tabular}{|c|c|c|c|}
\hline
\textbf{Epoch\_No.} & \textbf{Avg\_JSD} & \textbf{Avg\_WD} & \textbf{time\_stamp (s)}\\
\hline
0 & 0.19 & 0.08 & 24.26\\
\hline
1 & 0.082 & 0.04 & 48.46\\
\hline
...    & ...   & ... & ...  \\
\hline
499 & 0.01 & 0.003 & 12167.93\\
\hline
\end{tabular}
\label{tab:appendix_tab1}
\end{table}

\end{document}